\documentclass{article}
\usepackage{ijcai25}

\usepackage{times}
\usepackage{soul}
\usepackage{url}
\usepackage[hidelinks]{hyperref}
\usepackage[utf8]{inputenc}
\usepackage[small]{caption}
\usepackage{graphicx}
\usepackage{amsmath}
\usepackage{amsthm}
\usepackage{booktabs}
\usepackage{algorithm}
\usepackage{arydshln}
\usepackage{algorithmic}
\usepackage{amssymb}

\usepackage{subcaption}
\usepackage{mdframed}
\usepackage[switch]{lineno}
\usepackage{bm}
\usepackage{colortbl}
\usepackage{multirow}
\usepackage[table,xcdraw,dvipsnames]{xcolor}
\usepackage{threeparttable}
\definecolor{gray}{rgb}{0.3, 0.3, 0.3}
\definecolor{lightgray}{gray}{.9}

\definecolor{Q1blue}{RGB}{42,77,110}
\definecolor{Q2green}{RGB}{27,103,107}
\definecolor{Q3purple}{RGB}{74,35,90}
\newcolumntype{I}{!{\vrule width 1pt}}
\makeatletter
\newcommand{\thickhline}{%
    \noalign {\ifnum 0=`}\fi \hrule height 1pt
    \futurelet \reserved@a \@xhline
}

\newcommand{\redorangeup}[1]{$_{\color{RedOrange}\uparrow #1}$}
\makeatother

\newcommand{\fl}{\textsl{FL}}
\newcommand{\clip}{\textsl{CLIP}}

\newcommand{\cifarhun}{{Cifar-100}}

\newcommand{\officehome}{{Office-Home}}

\newcommand{\art}{{Art}}
\newcommand{\artabbrv}{{Ar}}
\newcommand{\clipart}{{Clipart}}
\newcommand{\clipartabbrv}{{C}}
\newcommand{\product}{{Product}}
\newcommand{\productabbrv}{{P}}
\newcommand{\realworld}{{Real World}}
\newcommand{\realworldabbrv}{{RW}}

\newcommand{\amazon}{{Amazon}}
\newcommand{\amazonabbrv}{{Am}}
\newcommand{\webcam}{{Webcam}}

\newcommand{\domainnet}{{DomainNet}}
\newcommand{\sketch}{{Sketch}}
\newcommand{\sketchabbrv}{{S}}
\newcommand{\webcamabbrv}{{W}}
\newcommand{\dslr}{{DSLR}}
\newcommand{\dslrabbrv}{{D}}
\newcommand{\realabbrv}{R}

\newcommand{\officeto}{{Office31}}

\newcommand{\coop}{{CoOP}}
\newcommand{\promptsrc}{{PromptSRC}}
\newcommand{\vpt}{{VPT}}

\usepackage[capitalize]{cleveref}
\crefname{section}{Sec.}{Secs.}
\crefname{table}{Tab.}{Tabs.}
\crefname{proposition}{Prop.}{Props.}

\title{An Empirical Study of Federated Prompt Learning for Vision Language Model}

\author{
Zhihao Wang$^{*1}$
\and
Wenke Huang$^{*1}$\and
Tian Chen$^{*1}$\and
Zekun Shi$^1$\and
Guancheng Wan$^1$\and
Yu Qiao$^1$\and \\
Bin Yang$^1$\and
Jian Wang$^{\dagger}$$^{1,2}$\and
Bing Li$^{\dagger}$$^{1,2}$\And
Mang Ye$^{\dagger}$$^1$
\\
\affiliations
$^1$School of Computer Science, Wuhan University\\
$^2$Zhongguancun Laboratory, China
\emails
\{zhihao\_wang, wenkehuang, tian.chen\}@whu.edu.cn
}

\begin{document}

\maketitle

\begin{abstract}
    The Vision Language Model (VLM) excels in aligning vision and language representations, and prompt learning has emerged as a key technique for adapting such models to downstream tasks. However, the application of prompt learning with VLM in federated learning (\fl{}) scenarios remains underexplored. This paper systematically investigates the behavioral differences between language prompt learning (LPT) and vision prompt learning (VPT) under data heterogeneity challenges, including label skew and domain shift. We conduct extensive experiments to evaluate the impact of various \fl{} and prompt configurations, such as client scale, aggregation strategies, and prompt length, to assess the robustness of Federated Prompt Learning (FPL). Furthermore, we explore strategies for enhancing prompt learning in complex scenarios where label skew and domain shift coexist, including leveraging both prompt types when computational resources allow. Our findings offer practical insights into optimizing prompt learning in federated settings, contributing to the broader deployment of VLMs in privacy-preserving environments.
\end{abstract}

\section{Introduction}
\label{sec:intro}
\begin{figure}[t]
	\centering
	\includegraphics[width=0.95\linewidth]{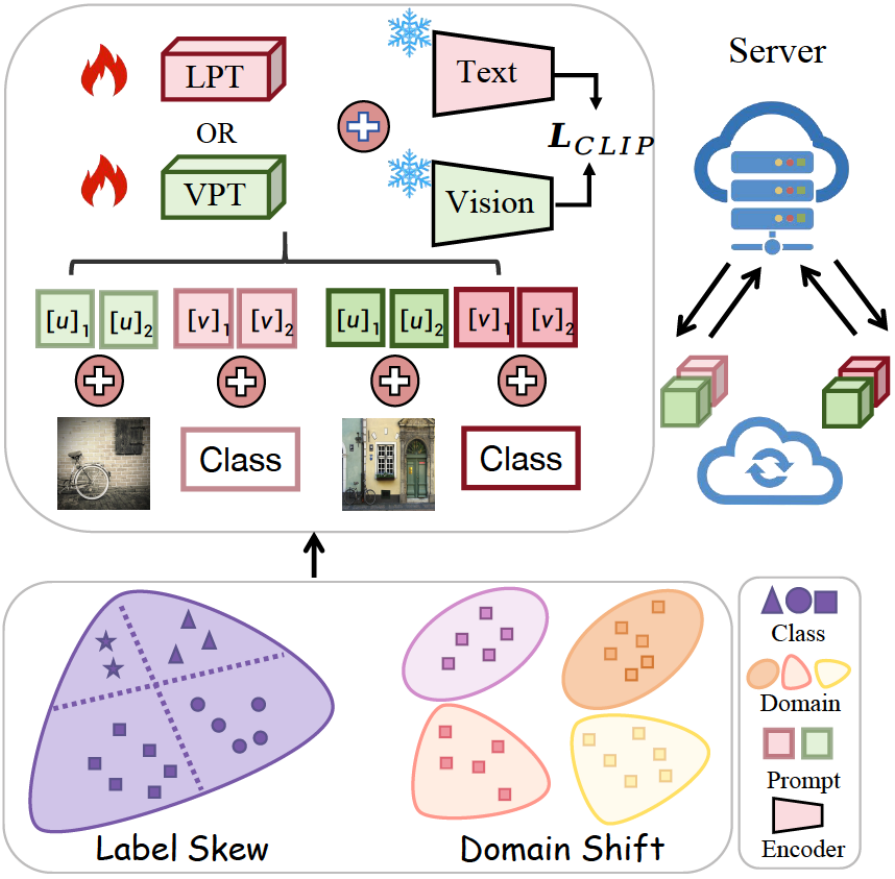}

	\captionsetup{font=small}
	\caption{
		\textbf{Background} for Federated Prompt Learning. Participants learn shared visual or textual prompts. However, under data heterogeneity (Label Skew and Domain Shift), aggregated prompts struggle to capture consistency among clients. See details in \cref{sec:intro}. } 
	\label{fig:problem}

\end{figure}

The advent of large-scale Vision Language Models, such as Contrastive Language-Image Pretraining (\clip{}) \cite{CLIP_21}, has redefined multi-modal learning by enabling effective alignment between visual and textual representations. These models have demonstrated remarkable adaptability across various downstream tasks with prompt tuning \cite{AutoPrompt_EMNLP20,Prompttuning_emnlp21}, particularly in zero-shot and few-shot settings. However, due to the inconsistency of upstream and downstream data distribution and training costs, large-scale Vision Language Models such as \clip{} still face many challenges in training and deployment.

\par
Prompt learning \cite{active_cvpr24,promptkd_CVPR24}, a lightweight yet effective paradigm for fine-tuning large pre-trained models, has emerged as a promising solution. By introducing task-specific prompts rather than retraining full model parameters, prompt learning offers an efficient means of adapting models to new tasks. While this approach has been extensively examined in centralized settings, its potential for optimizing VLM under federated conditions remains underexplored \cite{PromptFL_TMC23}. In these scenarios, the design and tuning of prompts are particularly critical, as they must account for the decentralized nature of data and the limited communication resources available in \fl{}.
\par
In Federated Learning scenarios, the challenges of data heterogeneity \cite{FLwithNonIID_arXiv18,Advances_arXiv19,FLChallengesMethodsDirection_SPM20,FedStar_AAAI23} are further compounded by issues like \textbf{Label Skew and Domain Shift} as shown in \cref{fig:problem}. Label skew occurs when different clients have significantly imbalanced class distributions, leading to poor generalization on unseen data. Domain shift arises when the underlying data distributions vary across clients, making it difficult for the model to converge to a globally optimal solution. For VLMs enhanced with prompt learning, these issues manifest in distinct ways: language prompts may be less effective when the semantic representation of classes varies widely due to label skew, whereas visual prompts may struggle to capture consistent image features across diverse domains.

\par
Motivated by these observations, our work conducts an in-depth empirical study of prompt learning for VLM in \fl{} settings.
We systematically analyze the distinct behaviors of language prompt learning and vision prompt learning in scenarios characterized by Label Skew and Domain Shift. Specifically, we explore how each type of prompt responds to imbalanced class distributions and varying domain characteristics, thereby uncovering the strengths and limitations inherent in each approach.
We study the influence of different \fl{} configurations alongside various prompt settings. This analysis aims to explore the robustness of federated prompt learning.
Furthermore, recognizing that real-world applications often involve simultaneous challenges from both Label Skew and Domain Shift, we investigate strategies to boost the effectiveness of prompt learning in such complex environments.
\par
Our investigation is structured around three main research objectives:
\begin{itemize}

  \item {\protect \includegraphics[scale=0.11]{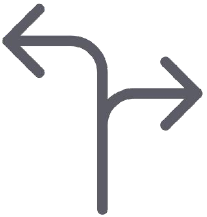}} \textbf{{\textcolor{Q1blue}{Q1:}}} \textit{Do Language Prompt Tuning (LPT) and Vision Prompt Tuning (VPT) show} \textbf{\textcolor{Q1blue}{Behavior Discrepancies}} \textit{under label skew and domain shift?}  
  
  \item {\protect \includegraphics[scale=0.08]{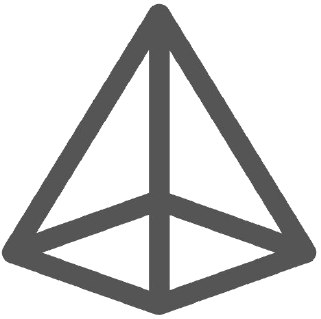}} \textbf{{\textcolor{Q2green}{Q2:}}} \textit{Does federated prompt learning exhibits} \textbf{\textcolor{Q2green}{Robustness Patterns}} \textit{in various federated settings and prompt configurations?}  
  
  \item {\protect\includegraphics[scale=0.25]{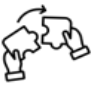}} \textbf{{\textcolor{Q3purple}{Q3:}}} \textit{Do federated visual and textual prompts appear} \textbf{\textcolor{Q3purple}{Collaboration Effect}} \textit{for complex heterogeneity?}  
  
\end{itemize}

\par
By addressing these three key areas, our work not only elucidates the differential impacts of language and vision prompts in federated settings but also provides practical guidelines for optimizing prompt learning in large-scale, distributed vision-language models. This study paves the way for more robust and resource-efficient deployments of VLM in real-world \fl{} applications.

\section{Related Work}
\label{sec:related}

\subsection{Federated Learning}
Federated Learning (\fl{}) enables collaborative model training across decentralized clients while preserving data privacy \cite{FedOptimization_16,Advances_arXiv19,Challenges_20,FCCL_22,Tackle_23,FCCLPlus_TPAMI23,FLSurveyandBenchmarkforGenRobFair_PAMI24}. However, it faces significant challenges, particularly in scenarios involving label skew and domain shift, which substantially degrade model performance due to the presence of non-IID (non-independent and identically distributed) data \cite{CCVR_NeurIPS21,Non-IIDStudy_22,SOTAIIDSurvey_22,Heterogeneous_Survey_2023,Tackle_23,FedMut_24,FREIB_AAAI2025}. Label skew occurs when the class distributions vary significantly across clients, leading to poor generalization on unseen data. Domain shift, on the other hand, arises when the input data distribution differs across clients, further complicating model convergence.
\par
To address label skew, FedProx \cite{FedProx_20}introduced a proximal term to the optimization objective, stabilizing training in non-IID settings. Similarly, MOON \cite{MOON_21} proposed contrastive learning at the client level to align model representations and mitigate discrepancies caused by skewed labels. For domain shift, FedBN \cite{FedBN_ICLR21} employed client-specific batch normalization layers, allowing local domain-specific feature extraction without compromising global model updates. Other methods, such as MFL\cite{MFL_20} and CFL\cite{CFL_21}, attempted to balance global and local objectives by clustering clients with similar data distributions or introducing momentum-based updates to improve convergence under domain shifts.
These methods focus on improving the communication efficiency and convergence of \fl{} systems when clients experience heterogeneous data distributions. Despite these advancements in handling label skew and domain shift individually, there is a need for further research on how large-scale models like \clip{} can be better adapted and applied in \fl{} settings when facing these challenges.  This paper aims to explore how \clip{} can leverage prompt learning to maintain robust performance in federated settings, ensuring better cross-client adaptation and efficient model convergence despite the inherent challenges of data distribution.

\subsection{Prompt Learning}
Prompt learning is a powerful and flexible approach for adapting pre-trained models to new tasks with minimal adjustments. Initially developed for NLP and widely used in LLMs like GPT-3 \cite{gpt3_nips20}, it has since extended to vision-language models. Techniques like prompt tuning \cite{Prompttuning_emnlp21} and prefix tuning show that large models can be efficiently adapted using lightweight prompt modifications without full retraining.
\par
In Vision Language Models like \clip{}, prompt learning has been particularly successful in enhancing the zero-shot and few-shot learning capabilities of the model \cite{active_cvpr24,promptkd_CVPR24}. Language prompts enable the model to better align textual and visual representations, improving performance on tasks like cross-modal retrieval. KgCoOp \cite{KgCoOp_cvpr23} enhances language prompts by reducing the discrepancy between learned and handcrafted prompts, thereby improving performance in unseen classes. Vision prompts, on the other hand, modify the image encoding process, allowing the model to adapt to visual-specific tasks without fine-tuning the entire network\cite{PLOT_ICLR23}. Besides, MaPLe \cite{MaPLe_cvpr23} introduces multi-modal prompt learning, combining language and vision prompt learning. These methods have shown promising results when applied to vision-language models, improving the model's performance on tasks such as zero-shot image classification and visual question answering. However, the methods above did not provide a comprehensive study of prompt learning in \fl{}. This paper explores both language and vision prompt learning of VLM in \fl{}, where data heterogeneity shows a strong influence on model performance.

\subsection{Federated Prompt Learning}
Recent research has explored integrating prompt learning into federated systems to collaboratively train shared prompts, enhancing the generalization ability for specific tasks {\cite{FedCLIP_DEB23,PromptFL_TMC23,FedAPT_AAAI24,FedDPT_arXiv23,SurveyofPromptVL_arXiv23}}. For instance, FedPR \cite{FedPR_CVPR23} leverages federated visual prompts within the null space, while {\cite{VPwithDP_ICCV23}} highlights the role of visual prompts in balancing the trade-off between privacy and utility while minimizing privacy budget consumption.
In addition to visual prompts, several studies have investigated textual modality adaptation. PromptFL \cite{PromptFL_TMC23} focuses on learning a shared text prompt, whereas FedTPG {\cite{FedTPG_ICLR24}} introduces a unified prompt generation framework to coordinate prompt learning across clients. Moreover, personalized prompt learning has gained attention, with recent works exploring client-specific prompt optimization strategies to address data heterogeneity \cite{pFedPG_ICCV23,pFedPrompt_WWWW23,FedAPT_AAAI24}. FedOTP \cite{FedOTP_CVPR24} proposes an efficient collaborative prompt learning approach that enables each client to capture distinct category characteristics individually. Nevertheless, these methods do not comprehensively test the performance of different prompt learning under various federated settings and prompt configurations.

\section{Method Details}
We briefly review the foundations of \clip{} and introduce the details of prompt learning of \clip{} in federated settings.

\subsection{Foundations of \clip{}}
Contrastive Language-Image Pretraining (\clip{}) revolutionized multi-modal representation learning by aligning visual and textual data through contrastive learning. Given a dataset of paired image-text examples $D=\left\{(I_k,T_k)\right\}_{k=1}^{|D|}$, the goal of \clip{} is to encourage semantic alignment between image and text embeddings while repelling unpaired samples. The architecture of \clip{} includes a visual encoder $f_v$ and a text encoder $f_t$,which transform an image $I_k$ and a corresponding text $T_k$ into normalized feature embeddings $\mathbf{z}_k=f_v(I_k)$ and $\mathbf{w}_k=f_t(T_k)$ , respectively.
\par
The training objective employs a symmetric contrastive loss based on the InfoNCE \cite{infonce_arXiv18} framework. For an image embedding $\mathbf{z}_k=f_v(I_k)$, the image-to-text contrastive loss is defined as:

\begin{equation}
    \mathcal{L}_{I\to T}(\mathbf{z},\mathbf{w})=-\log\frac{\exp(\mathbf{z}_k\cdot \mathbf{w}_k /\tau)}{\sum_{b=1}^{|B|}\exp(\mathbf{z}_k\cdot \mathbf{w}_b/\tau)},
\end{equation}
\noindent where $\tau$ is a learnable temperature parameter controlling distribution scaling, 
$\cdot$ denotes the dot product, and $B$ represents the batch size.
\par
Similarly, the text-to-image contrastive loss, anchored on a text embedding 
$\mathbf{z}_k$, is formulated as:

\begin{equation}
        \mathcal{L}_{T\to I}(\mathbf{z},\mathbf{w})=-\log\frac{\exp(\mathbf{w}_k\cdot \mathbf{z}_k /\tau)}{\sum_{b=1}^{|B|}\exp(\mathbf{w}_k\cdot \mathbf{z}_b/\tau)}.
\end{equation}

\par
The overall CLIP loss combines these two components symmetrically:
\begin{equation}
    \mathcal{L}(\mathbf{z},\mathbf{w})=\frac{1}{2}(\mathcal{L}_{I\to T}+\mathcal{L}_{T\to I})
\end{equation}

\subsection{Prompt learning of \clip{} in Federated Learning}
In this section, we describe how both language and vision prompt learning techniques are adapted for \clip{} in \fl{}.

\subsubsection{Language Prompt}
In Language Prompt Learning (LPT), we utilize a learnable vector $\mathbf{P}_t$, which is combined with the class token for the text input, following the approach proposed in \cite{CoOp_IJCV22}. Specifically, the text input $T_k$ is augmented with a series of prompt vectors $\{v_1, v_2, \cdots, v_L\}$ that are learned during training, where $L$ is the number of tokens in the prompt. The augmented text input to the \clip{} text encoder is:
\begin{equation}
	\Tilde{T}_k = [v_1, v_2, \cdots, v_L,\texttt{CLASS}],
\end{equation}
\noindent where \texttt{CLASS} is a fixed token that represents the class information (or task-specific information). The input $\Tilde{T}_k$ is then passed through the text encoder $f_t$, resulting in the modified text embedding $\Tilde{\mathbf{w}}_t $:
\begin{equation}
	\Tilde{\mathbf{w}}_k=f_t(\Tilde{T}_k).
\end{equation}
The $\Tilde{\mathbf{w}}_k$ is normalized before being used in downstream tasks. The use of learnable text prompts allows the model to adapt to domain-specific language and task-specific variations, improving performance in tasks where text data is heterogeneous across clients.

\subsubsection{Vision Prompt}
For Vision Prompt Learning (VPT), we follow the method in \cite{VPT_ECCV22}, where a series of learnable vectors $\{u_1, u_2, \cdots, u_L\}$ are inserted between the class token \texttt{CLS} and the image patch embeddings $\mathbf{E}$ for the image encoder. The vision prompt $\mathbf{P}_v$  is thus represented as a sequence of these learnable vectors, which are concatenated with the image embeddings. The input to the image encoder is:

\begin{equation}
	 \Tilde{I}_k= [\mathtt{CLS}, u_1, u_2 \cdots, u_L, \mathbf{E}].
\end{equation}
The modified image $\Tilde{I}_k$ is passed through the visual encoder $f_v$,  yielding the image embedding $\Tilde{\mathbf{z}}_k$:
\begin{equation}
	\Tilde{\mathbf{z}}_k=f_v(\Tilde{I}_k).
\end{equation}
Similar to the text embeddings, the resulting image embedding $\Tilde{\mathbf{z}}_k$ is normalized before being used in downstream tasks. This modification helps the model focus on relevant features and adapt to the visual characteristics specific to each client’s data distribution.
\subsubsection{Federated Prompt Learning}
In Federated Prompt Learning, the clients collaboratively learn shared prompt modules while maintaining the privacy of their local data. Following typical Federated Learning setup {\cite{FedAvg_AISTATS17,FedProx_20,MOON_21,FPL_CVPR23}} , we assume that there are $M$ clients (indexed by $M$). Each client $m$ holds its own dataset ${D}^m$ and optimizes its local prompts $\mathbf{P}^m$ ($\mathbf{P}_t$ or $\mathbf{P}_v$). The goal is to aggregate the local prompts across clients into a global prompt $\mathbf{P}^g$, which is then used to update the global model. The process can be divided into three steps: distribution, optimization, and aggregation.
%\begin{equation}
%	\begin{aligned}
%		\mathbf{P}^m \leftarrow{}\mathbf{P}^g  & &  \text{Distribution},\\
%		\text{arg}\min_{\mathbf{P}^m}\mathbb{E}_{(x, {y})\sim D^m} \, \mathcal{L}(\mathbf{\Tilde{z}},\mathbf{\Tilde{w}})  & & \text{Optimization},\\
%		\mathbf{P}^g = \sum_m^M  \frac{N^m}{N} \mathbf{P}^m & & \text{Aggregation}.\\
%	\end{aligned}
%	\label{eq:federatedpfl}
%\end{equation}

%\begin{equation}
%	\begin{aligned}
%		\mathbf{P}^m \leftarrow{}\mathbf{P}^g  & &  \text{Distribution},\\
%		\begin{cases}
%			\text{arg}\min_{\mathbf{P}^m}\mathbb{E}_{(x, {y})\sim D^m} \, \mathcal{L}(\mathbf{{z}},\mathbf{\Tilde{w}}), \text{LPT}\\
%			\text{arg}\min_{\mathbf{P}^m}\mathbb{E}_{(x, {y})\sim D^m} \, \mathcal{L}(\mathbf{\Tilde{z}},\mathbf{{w}}), \text{VPT}
%		\end{cases}  & & \text{Optimization},\\
%		\mathbf{P}^g = \sum_m^M  \frac{N^m}{N} \mathbf{P}^m & & \text{Aggregation},\\
%	\end{aligned}
%\end{equation}
\begin{equation}
	\begin{aligned}
		\mathbf{P}^m \leftarrow{}\mathbf{P}^g  & &  \text{Distribution},\\
			\text{arg}\min_{\mathbf{P}^m}\mathbb{E}_{(x, {y})\sim D^m} \, \mathcal{L}(\mathbf{{z}},\mathbf{\Tilde{w}})/ \mathcal{L}(\mathbf{\Tilde{z}},\mathbf{{w}})
 & & \text{Optimization},\\
		\mathbf{P}^g = \sum_m^M  \frac{N^m}{N} \mathbf{P}^m & & \text{Aggregation},\\
	\end{aligned}
		\label{eq:federatedpfl}
\end{equation}
\noindent where $N^m$ is the number of samples at client $m$, and $N=\sum_m^M N^m$ is the total number of samples across all clients.
The global prompt $\mathbf{P}^g$ is then sent back to all clients, and the process is repeated iteratively. Through this federated learning setup, clients collaboratively adapt the shared prompts to their local data distributions, while the global model maintains privacy and benefits from the collective knowledge.
\section{Experiments Setup}

\subsection{Datasets}
Following \cite{MOON_21,FPL_CVPR23}, we comprehensively evaluate the federated prompt learning with \clip{} on the following four datasets. 

\begin{itemize}
	\item \textbf{\cifarhun{}} \cite{cifar_Toronto09} is a famous  classification dataset, containing 32 $\times$ 32 images of 100 classes. Training and validating sets are composed of $50,\!000$  and $10,\!000$ images. 
	\item \textbf{\domainnet{}} \cite{DomainNet_ICCV19} contains $6$ domains: \sketch{} (\sketchabbrv), real (\realabbrv), painting (P),  infograph (I), quickdraw (Q), and clipart (\clipartabbrv) with $365$ classes. It consists $48k-172k$ images per domain.
	\item \textbf{\officehome} \cite{officehome_CVPR17} is a large classification dataset with $65$ classes and includes four different domains: \art{} (\artabbrv{}), \clipart{} (\clipartabbrv), \product{} (\productabbrv), and \realworld{} (\realworldabbrv{}).
	\item \textbf{\officeto{}} \cite{office31_ECCV10} has 31 classification number in three domains: \amazon{} (\amazonabbrv), \dslr{} (\dslrabbrv{}), and \webcam{} (\webcamabbrv{}). It consists of common objects in office scenarios, such as laptops, keyboards, and, file cabinets. 
\end{itemize}
\subsection{Implementation Details}
We present the experimental setup from three key aspects as follows:
\begin{itemize}
	\item {Network Structure}: We employ the pre-trained CLIP model \cite{CLIP_21} with a ViT-B/16 image encoder backbone, using the official implementation from the repository\footnote{https://github.com/openai/CLIP}. For prompt construction, both for the text and visual encoders, we refer to the open-source implementations of \coop{}\cite{CoOp_IJCV22}, \vpt{}\cite{VPT_ECCV22}, and \promptsrc{}~\cite{PromptSRC_ICCV23} available at\footnote{https://github.com/KaiyangZhou/CoOp}$^{,}$\footnote{https://github.com/KMnP/vpt}$^{,}$\footnote{https://github.com/muzairkhattak/PromptSRC}. The prompt length for both visual and textual prompts is set to $L = 16$. Visual prompts $\mathbf{P}_v$ are inserted into the first 9 layers of the transformer. Both textual and visual prompts are randomly initialized with the normal distribution.
	\item Client Scale Construction: For datasets \officeto{} \cite{office31_ECCV10}, \officehome{} \cite{officehome_CVPR17}, and \domainnet{} \cite{DomainNet_ICCV19}, which include multiple domains, we design the number of participants to be twice the number of domains. For example, we set the participant size to $K = 6$ for \officeto{}. For \cifarhun{}, we set the client scale to $K = 10$.
	\item Training Settings: To ensure a fair comparison, we follow \cite{FPL_CVPR23} and set the communication epoch $T = 50$ and the local update round $E = 1$ for all settings. The training batch size is $32$, and we use SGD as the local update optimizer. The corresponding weight decay is $\eta = 1e-5$ and momentum is set to $0.9$. The local client learning rate is $0.001$ for all scenarios. To mimic the label skew, we follow the common experiments setting and use Dirichlet sampling \cite{Dirichlet}. We fix the random seed at $0$ to ensure reproducibility and run the experiments on NVIDIA 3090.
\end{itemize}
\section{Results and Analysis}
Based on the above settings, we conduct extensive experiments to explore different types of prompt learning with \clip{} under federated learning settings in detail.

%%%%%%%%%% Label skew 大表 %%%%%%%%%%
\begin{table}[t]\small
\centering
\scriptsize{
\resizebox{\linewidth}{!}{
		\renewcommand\arraystretch{1.1}
\begin{tabular}{r||cccc}
\hline\thickhline
\rowcolor{lightgray}
& \multicolumn{4}{c}{Cifar100}\\
\cline{2-5}
\rowcolor{lightgray}
% \rowcolor{gray!10}
\multirow{-2}{*}{Methods} & $\beta=0.3$ & $\beta=0.5$ & $\beta=1.0$ & $\beta=5.0$\\

\hline\hline
% EFEFED C9C7C3 EAEBED
\rowcolor{gray!5}
ZS-CLIP
% & 64.88 & 64.88 & 64.88 & 64.88 \\
& \multicolumn{4}{c}{64.88} \\
\rowcolor[HTML]{D7F6FF}
VPT
& 78.74\redorangeup{13.86} & 79.44\redorangeup{14.56} & 79.38\redorangeup{14.50} & 79.76\redorangeup{14.88}\\
LPT
& 75.24\redorangeup{10.36} & 75.13\redorangeup{10.25} & 76.05\redorangeup{11.17} & 76.53\redorangeup{11.65}

\end{tabular}}}

\captionsetup{font=small}
\caption{\small{
\textbf{Accuracy of different prompts learning in Label Skew} scenarios. Please refer to \cref{sec:difference}.
}}
\label{tab:comare_label skew}

\end{table}

%%%%%%%%%% Domain 大表 %%%%%%%%%%
\begin{table*}[t]
\centering
\scriptsize{
\resizebox{\linewidth}{!}{
		\renewcommand\arraystretch{1.1}
\begin{tabular}{r||ccccIcIcccIcIccccccIc}
\hline\thickhline
\rowcolor{lightgray}
& \multicolumn{5}{cI}{Office-Home} &\multicolumn{4}{cI}{Office31} &\multicolumn{7}{c}{Domainnet}\\
\cline{2-17}
\rowcolor{lightgray}
\multirow{-2}{*}{Methods} & \textsl{A}  & \textsl{C}  & \textsl{PR} & \textsl{RW} & \textsl{AVG}
& \textsl{AM} & \textsl{D} & \textsl{W} & \textsl{AVG} & \textsl{C} & \textsl{I} &\textsl{P} &\textsl{Q} &\textsl{R} &\textsl{S} & \textsl{AVG}  \\

\hline\hline
\rowcolor{gray!5}
ZS-CLIP
& 84.30 & 66.17 & 89.18 & 89.66 & 82.32
& 80.96 & 72.45 & 74.05 & 75.82 
& 87.69 & 69.66 & 79.77 & 28.11 & 91.91 & 84.82 & 73.66 \\
VPT
& 84.13 & 75.23 & 93.84 & 92.14 & 86.34
& 87.26 & 90.00 & 92.53 & 89.93 
& 88.93 & 75.34 & 82.57 & 48.83 & 93.45 & 87.44 & 79.43 \\

\rowcolor[HTML]{D7F6FF}
LPT
& 86.20 & 75.87 & 94.57 & 93.70 & \textbf{87.58}\redorangeup{\textbf{5.26}}
& 89.40 & 92.25 & 95.32 & \textbf{92.32}\redorangeup{\textbf{16.50}}
& 90.30 & 75.84 & 84.05 & 44.59 & 93.24 & 87.58 & \textbf{79.27}\redorangeup{\textbf{5.61}}\\

\end{tabular}}}

\captionsetup{font=small}
\caption{\small{
\textbf{Accuracy of different prompts learning in Domain Shift} scenarios. See details in \cref{sec:difference}
}}
\label{tab:compare_domain}

\end{table*}

%%%%%%%%%% global sim 图 %%%%%%%%%%
\begin {figure*}[t]
\centering

\begin{minipage}{0.245\linewidth}
		\begin{subfigure}[b]{\textwidth}
		\includegraphics[width=\textwidth]{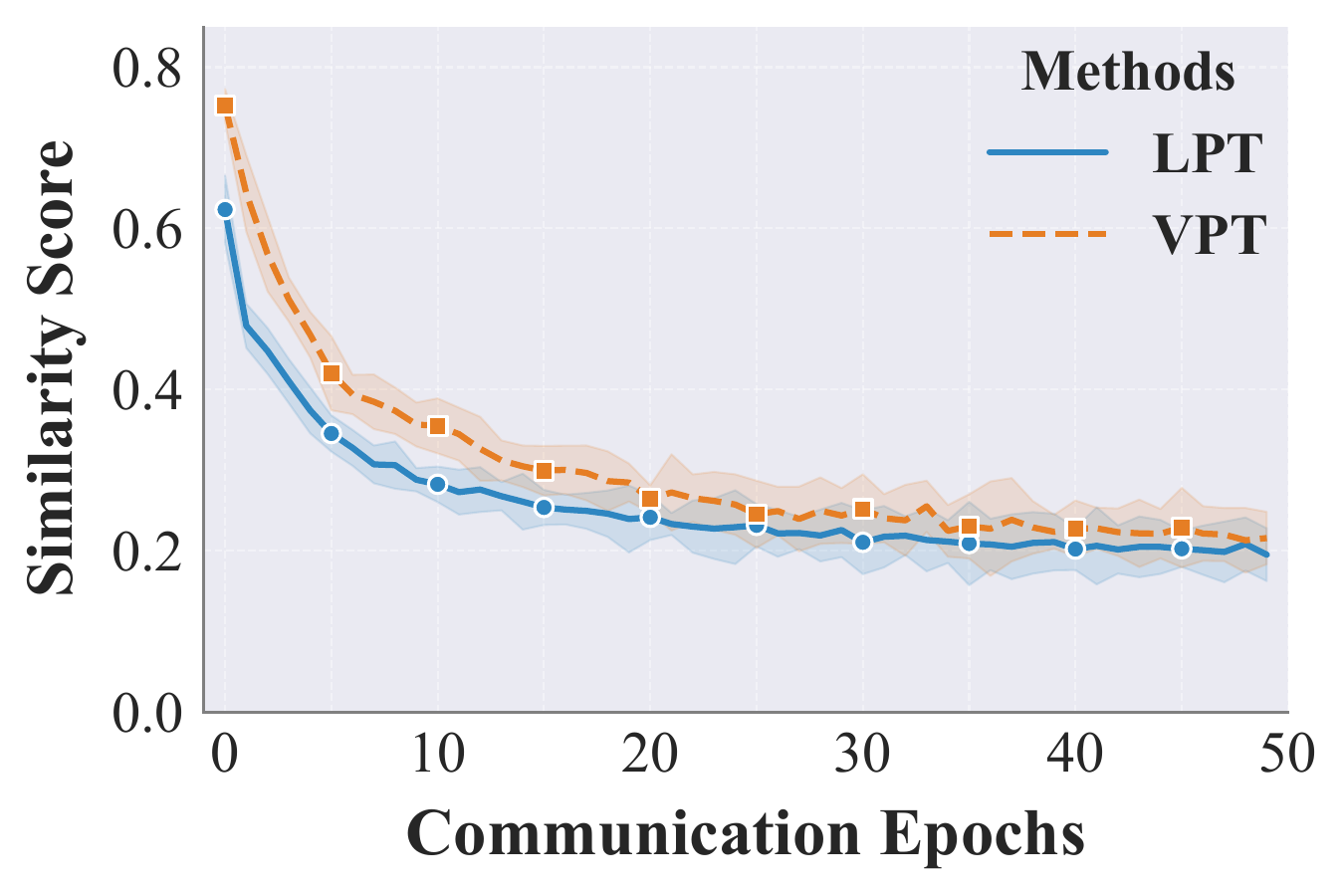}
				\caption{Mean Value of \cifarhun{}}
		\label{fig:globalcifarmean}
	\end{subfigure}

\end{minipage}
\begin{minipage}{0.245\linewidth}
		\begin{subfigure}[b]{\textwidth}
	\includegraphics[width=\textwidth]{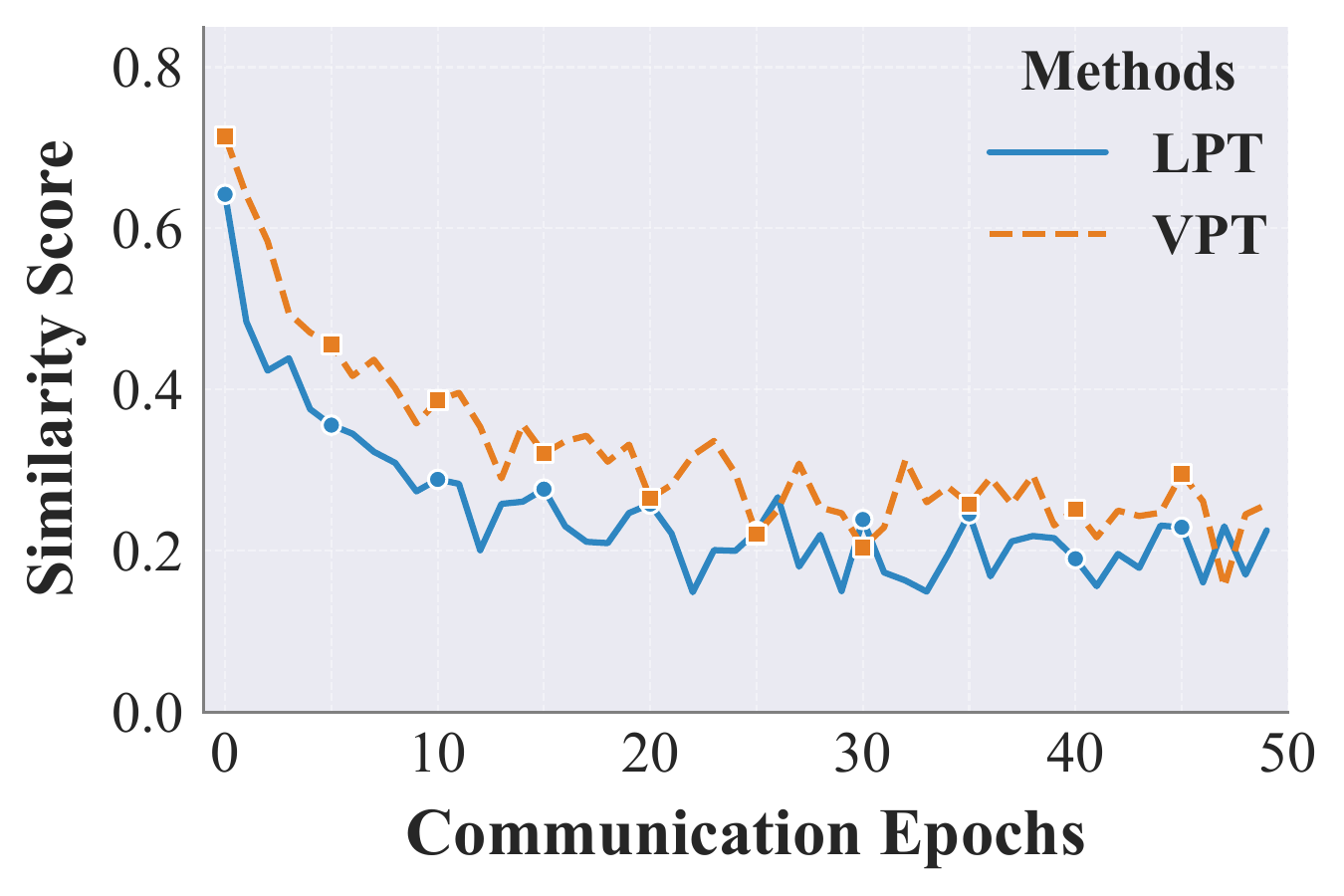}
	\caption{Single Client of \cifarhun{}}
	\label{fig:globalcifarclient}
\end{subfigure}

\end{minipage}
\begin{minipage}{0.245\linewidth}
		\begin{subfigure}[b]{\textwidth}
	\includegraphics[width=\textwidth]{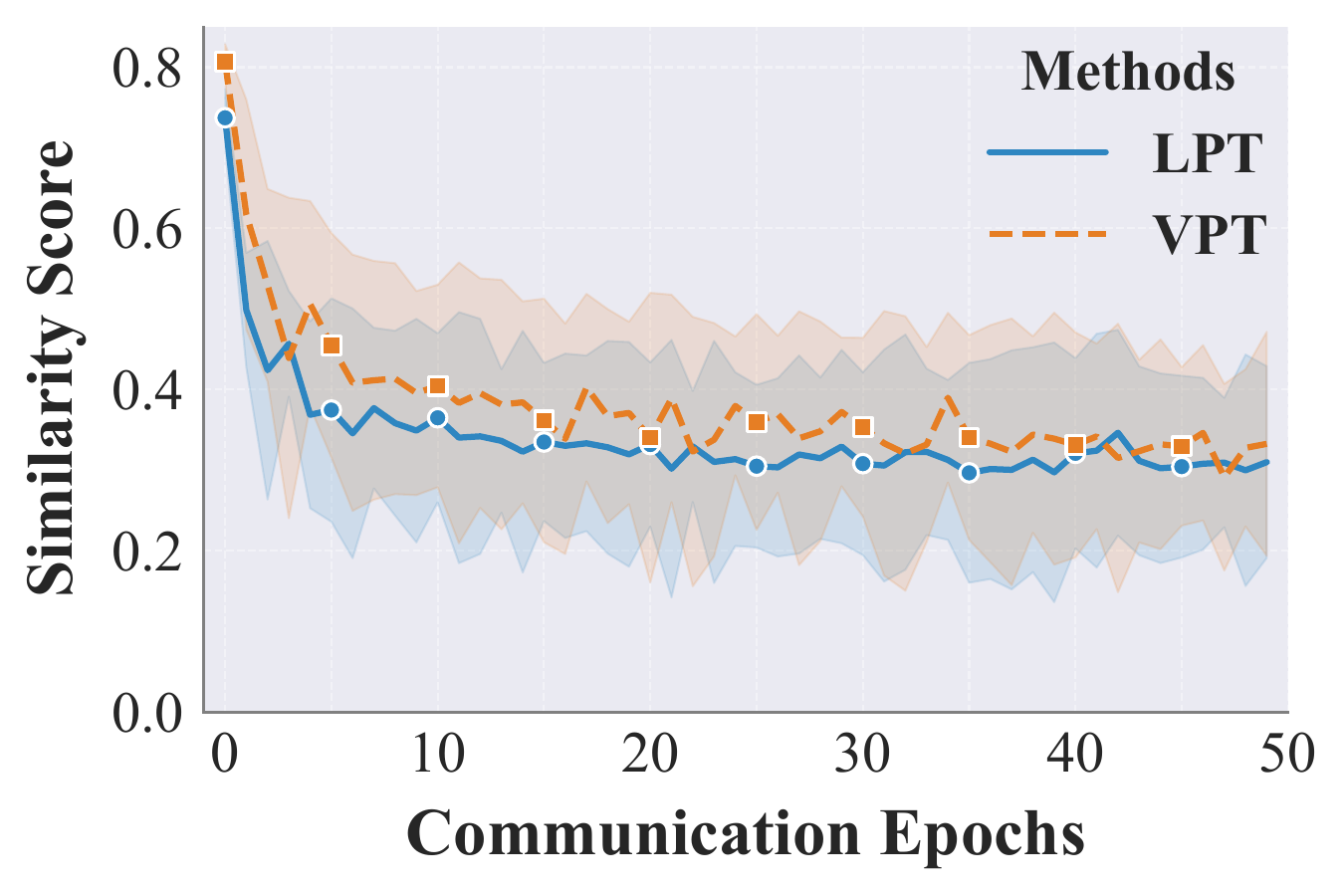}
	\caption{Mean Value of \officehome{}}
	\label{fig:globalofficemean}
\end{subfigure}

\end{minipage}
\begin{minipage}{0.245\linewidth}
		\begin{subfigure}[b]{\textwidth}
	\includegraphics[width=\textwidth]{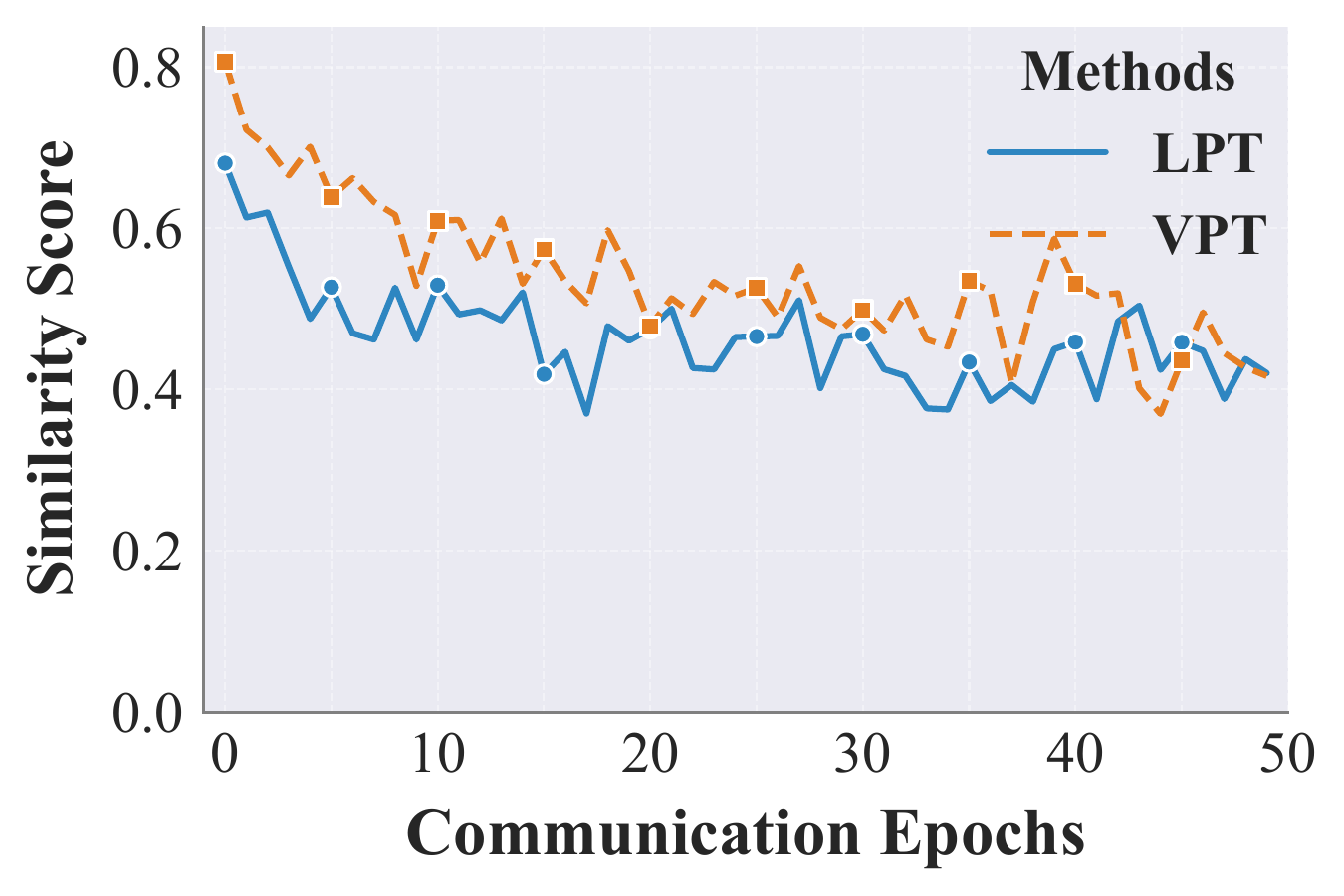}
	\caption{Single Client of \officehome{}}
	\label{fig:globalofficeclient}
\end{subfigure}

\end{minipage}
\captionsetup{}

\caption{\textbf{Similarities of Optimization Directions between clients and global server} in \cifarhun{} and \officehome{} in the training. \cref{fig:globalcifarmean} and \cref{fig:globalofficemean} illustrate the mean value of the similarity of optimization direction between all clients and the global server, while the shadows indicate the standard deviation. \cref{fig:globalcifarclient} and \cref{fig:globalofficeclient} are the selected clients. Please refer to \cref{sec:difference}.}

\label{fig:globalsim}

\end{figure*}

%%%%%%%%%% rounds参数 %%%%%%%%%%
\begin {figure}[htbp]
	\centering
	\includegraphics[width=1.0\linewidth]{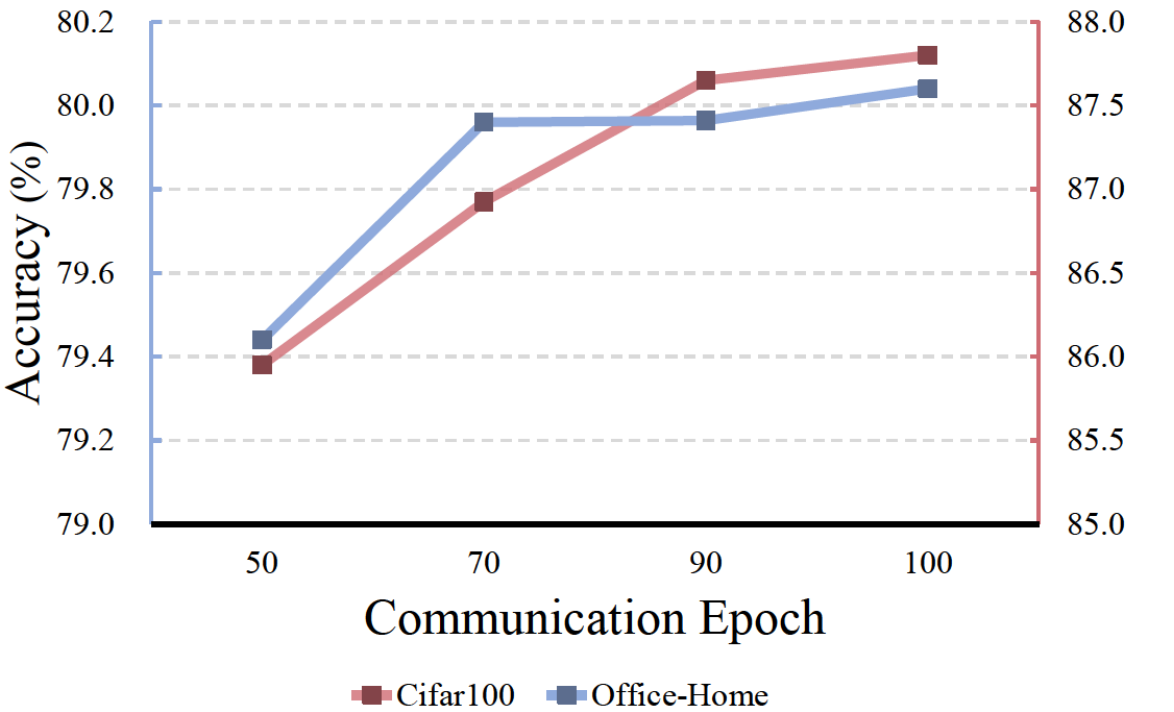}

	\captionsetup{}
 
	\caption{Effects of different \textbf{Communication Epochs}. Please refer to \cref{sec:robustness} }

	\label{fig:settingrounds}

\end{figure}

%%%%%%%%%% local sim 图 %%%%%%%%%%
\begin {figure*}[htbp]
\centering

\begin{minipage}{0.245\linewidth}
		\begin{subfigure}[b]{\textwidth}
		\includegraphics[width=\textwidth]{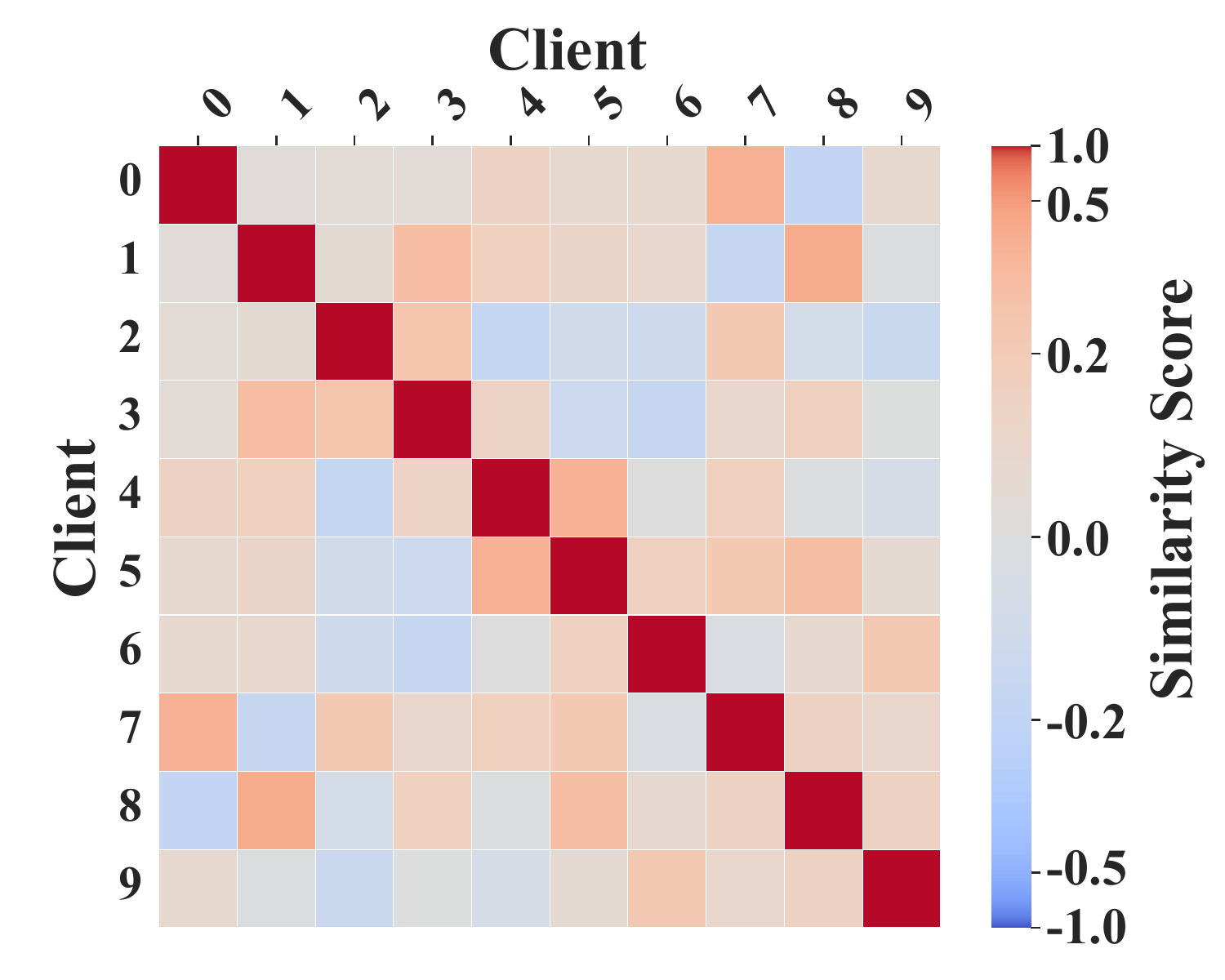}
				\caption{VPT in \cifarhun{}}
		\label{fig:localcifarvpt}
	\end{subfigure}

\end{minipage}
\begin{minipage}{0.245\linewidth}
		\begin{subfigure}[b]{\textwidth}
	\includegraphics[width=\textwidth]{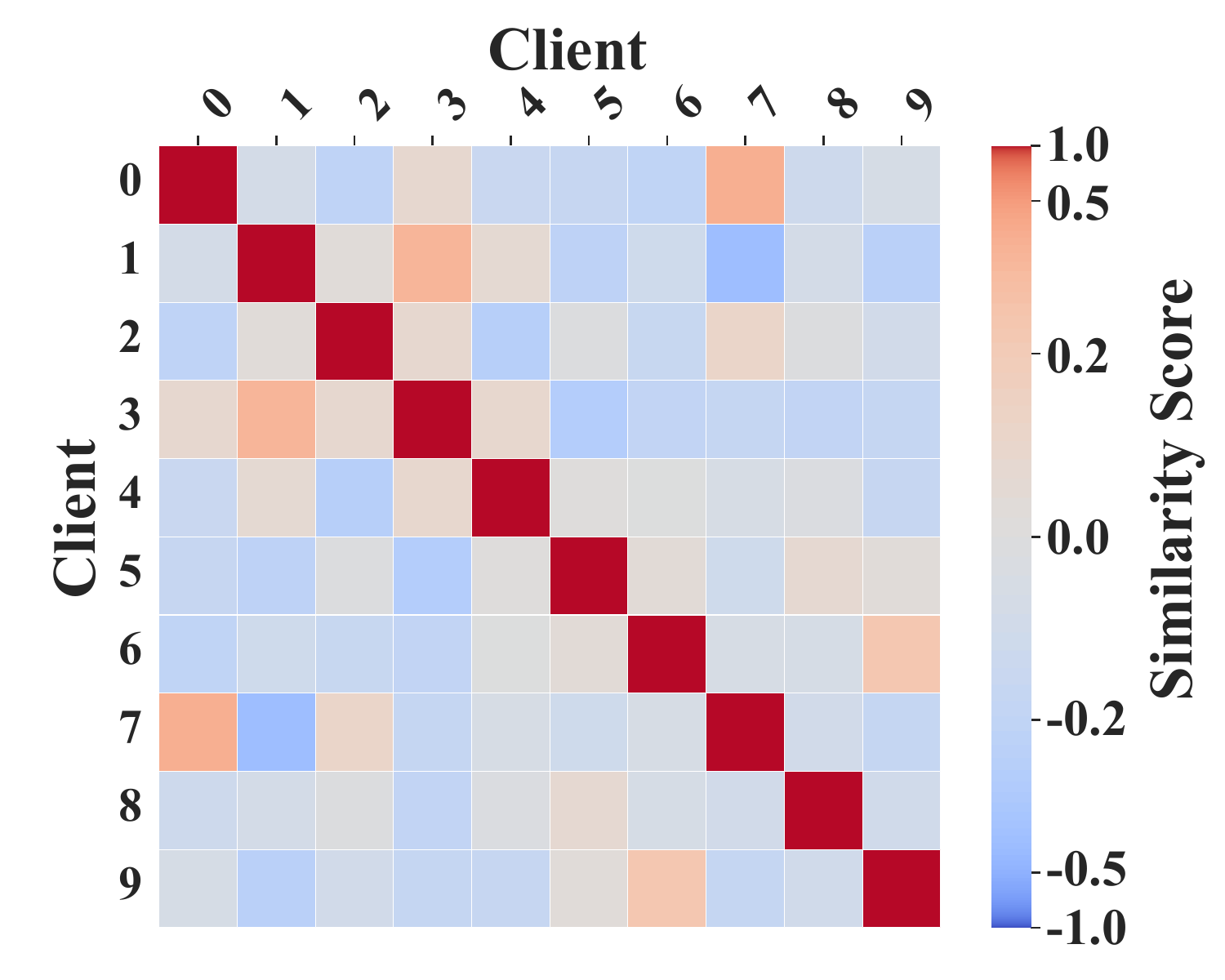}
	\caption{LPT in \cifarhun{}}
	\label{fig:localcifarlpt}
\end{subfigure}

\end{minipage}
\begin{minipage}{0.245\linewidth}
		\begin{subfigure}[b]{\textwidth}
	\includegraphics[width=\textwidth]{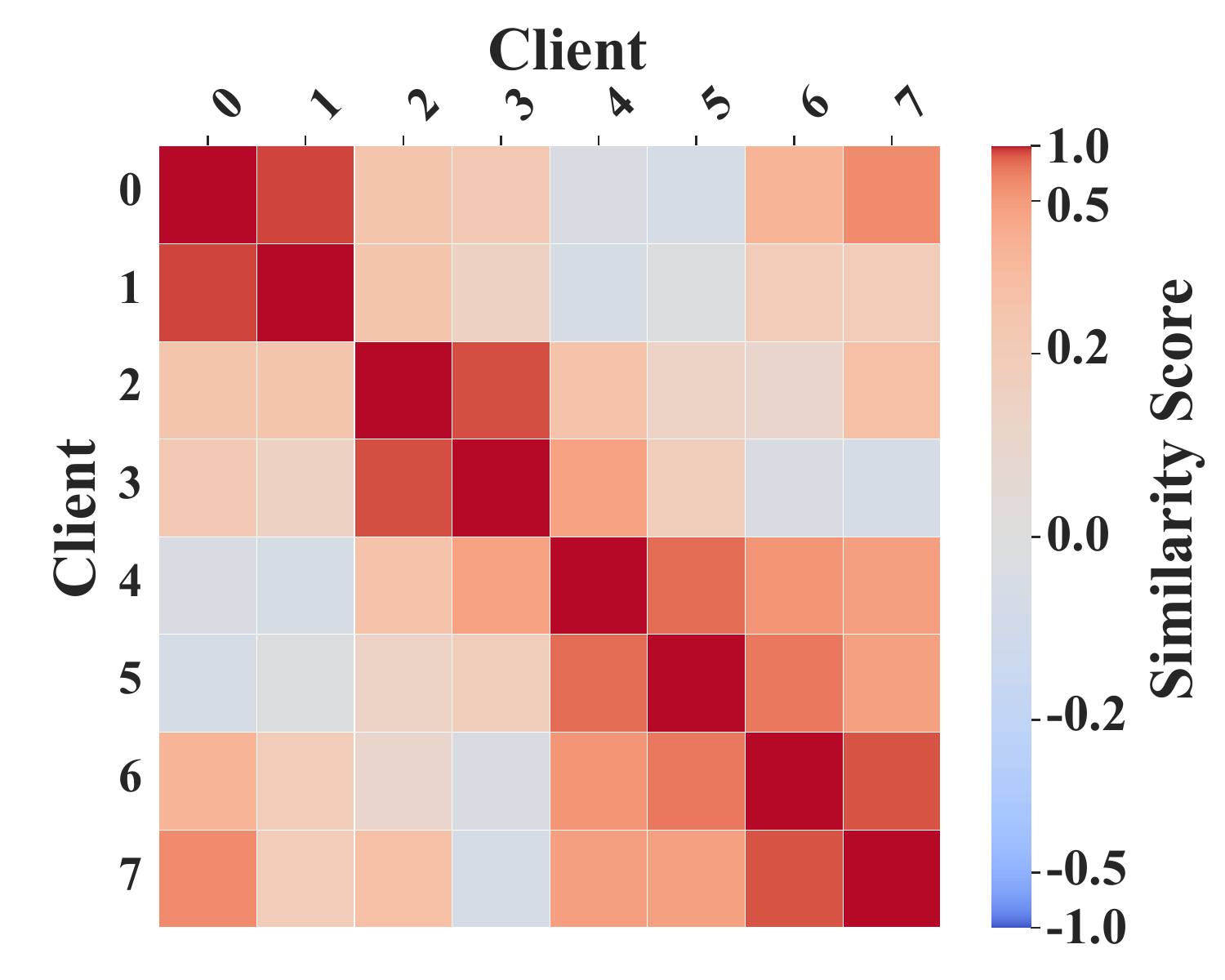}
	\caption{VPT in \officehome{}}
	\label{fig:localofficevpt}
\end{subfigure}

\end{minipage}
\begin{minipage}{0.245\linewidth}
		\begin{subfigure}[b]{\textwidth}
	\includegraphics[width=\textwidth]{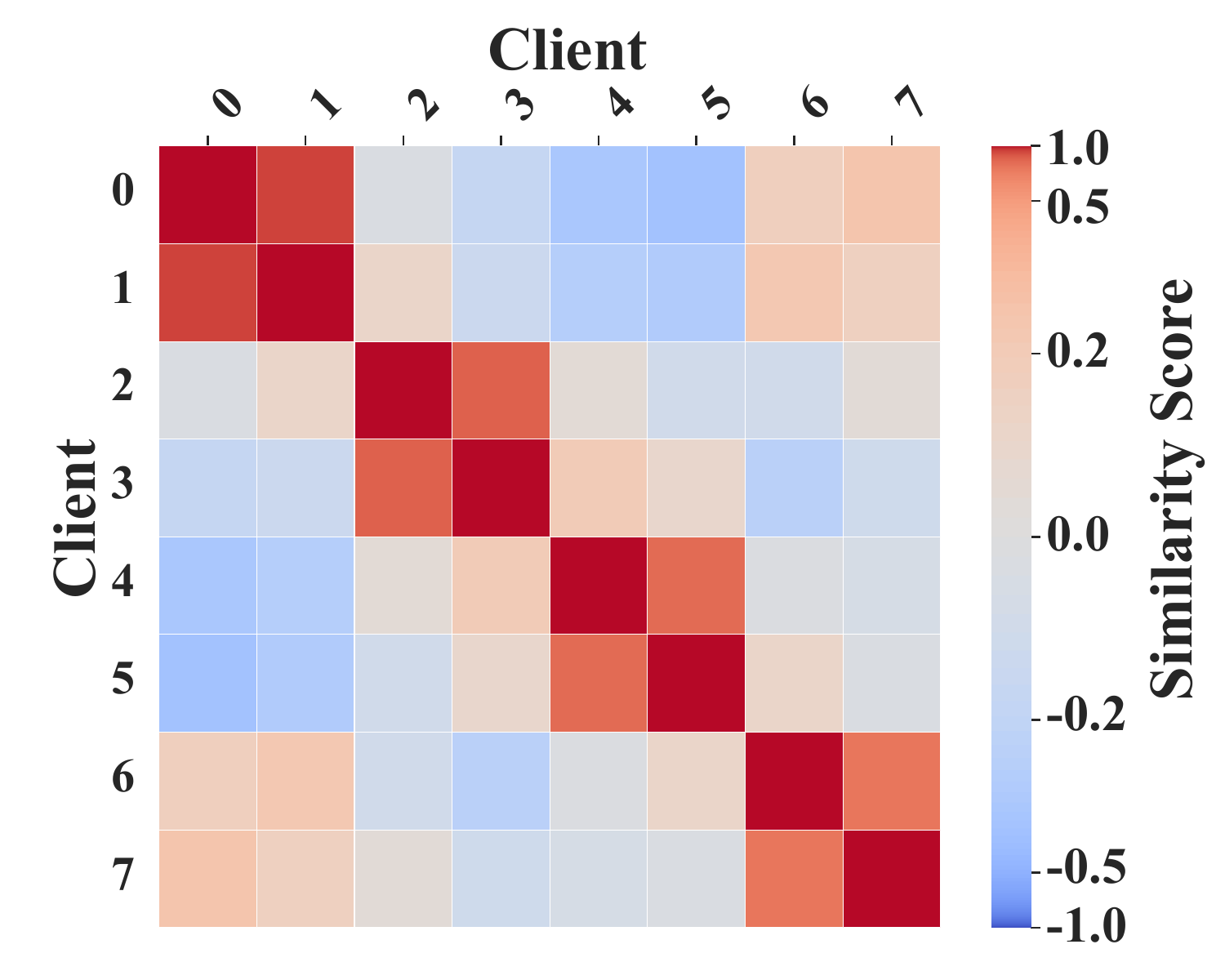}
	\caption{LPT in \officehome{}}
	\label{fig:localofficelpt}
\end{subfigure}

\end{minipage}
\captionsetup{}

\caption{\textbf{Similarities of Optimization Directions among clients} in \cifarhun{} and \officehome{} in the training epoch. \cref{fig:localcifarvpt} and \cref{fig:localcifarlpt} are the similarity between clients of VPT and LPT in \cifarhun{} respectively, while \cref{fig:localofficevpt} and \cref{fig:localofficelpt} stand for \officehome{} similarly. See details in \cref{sec:difference}.}

\label{fig:localsim}

\end{figure*}
	
\subsection{Behavior Discrepancy between LPT and VPT in Label Skew and Domain Shift}
\label{sec:difference}
To fully explore Language Prompt Learning (LPT) and Vision Prompt Learning (VPT) with \clip{} in federated learning settings, we conduct both experiments in the Label Skew and Domain Shift scenarios.
\par 
\noindent \textbf{Label Shift Scenarios.} Applying Dirichlet sampling, we evaluate VPT and LPT in \cifarhun{} with different degrees of Label Skew and the results are shown in \cref{tab:comare_label skew}. VPT outperforms LPT obviously in all settings.
\par
\noindent \textbf{Domain Shift Scenarios.} We conduct experiments on \officehome{}, \officeto{} and \domainnet{} as for Domain Shift scenarios. The results in \cref{tab:compare_domain} indicate that LPT works better than VPT in most domains.

\par
Since the results in the two scenarios are opposite, the key problem arises: \textit{What makes the differences of results in Label Skew and Domain Shift?} 
\par 
To address this problem, we further step in the training process in both scenarios. In concrete, we calculate the similarities of optimization directions of prompt learning between clients and the global server during training, and the visualization results are shown in \cref{fig:globalsim}. We can observe that in both Label Skew and Domain Shift scenarios, clients tend to optimize more similarly with the global server during the training of VPT. Besides, we also record the similarities among clients in \cifarhun{} and \officehome{}. We select the results of a single communication epoch and visualize them in \cref{fig:localsim}. Likewise, the optimization directions among clients are more similar in VPT than in LPT.
\par
Our experiments reveal that under label skew conditions, VPT exhibits significantly higher consistency in optimization directions among clients and between clients and the global server. This consistency helps align the visual features across disparate client data, effectively mapping the textual cues into a unified visual domain. In contrast, LPT shows more variability in its optimization paths, leading to inconsistent mapping and thus poorer performance when label distributions are imbalanced.
Under domain shift scenarios, the optimization directions in LPT are less similar across clients, which appears to allow each client to better adapt to its own domain-specific data. This individualized adaptation enables LPT to map the textual features into diverse visual domains more effectively, yielding superior performance. Conversely, the high consistency of VPT in optimization directions tends to ignore the unique characteristics of each domain, thereby limiting its adaptability and resulting in lower performance in heterogeneous domain settings. 
\par
Regarding \textbf{\textcolor{Q1blue}{Q1}} in \cref{sec:intro}, we can obtain the answer: \textbf{\textcolor{Q1blue}{A1}}: \textit{There exists \textbf{Behavior Discrepancy} between LPT and VPT under label skew and domain shift.}
We restate this key phenomenon as follows:
\definecolor{cadetblue}{RGB}{95,158,160} % Define CadetBlue color
\definecolor{keywordcolor}{RGB}{178,34,34} % Define FireBrick color for keywords
\begin{mdframed}[backgroundcolor=cadetblue!10, linewidth=0.8pt, linecolor=cadetblue!80, roundcorner=5pt]
\textbf{\textcolor{keywordcolor}{Label Skew Prompt Consistency}}:

\textit{For label skew, VPT exhibits consistent optimization across clients and with the global server, leading to unified visual mapping.}

\end{mdframed}

\begin{mdframed}[backgroundcolor=cadetblue!10, linewidth=0.8pt, linecolor=cadetblue!80, roundcorner=5pt]
\textbf{\textcolor{keywordcolor}{Domain Shift Prompt Discrepancy}}:

\textit{For domain shift, diverse optimization in LPT enables better adaptation to distinct domains. }
\end{mdframed}

\subsection{Robustness of Federated Prompt Learning Under Various Settings}
\label{sec:robustness}

%%%%%%%%%% aggregation参数 %%%%%%%%%%
\begin {figure}[htbp]
	\centering
	\begin{minipage}{0.49\linewidth}
		\begin{subfigure}[b]{\textwidth}
		\includegraphics[width=\textwidth]{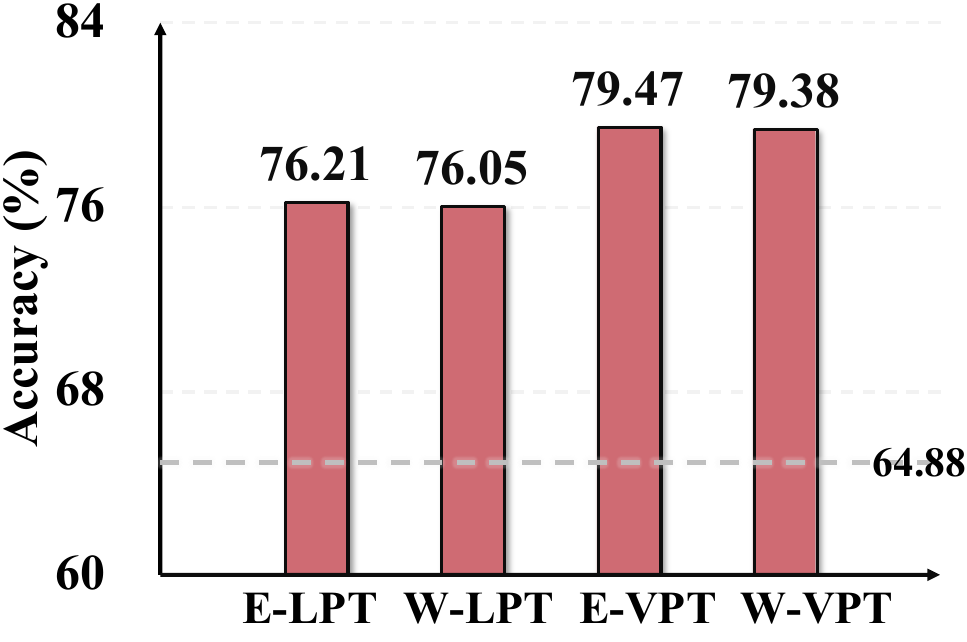}
				\caption{ \cifarhun{}}
		\label{fig:cifaragg}
	\end{subfigure}
    \end{minipage}
	\begin{minipage}{0.49\linewidth}
		\begin{subfigure}[b]{\textwidth}
		\includegraphics[width=\textwidth]{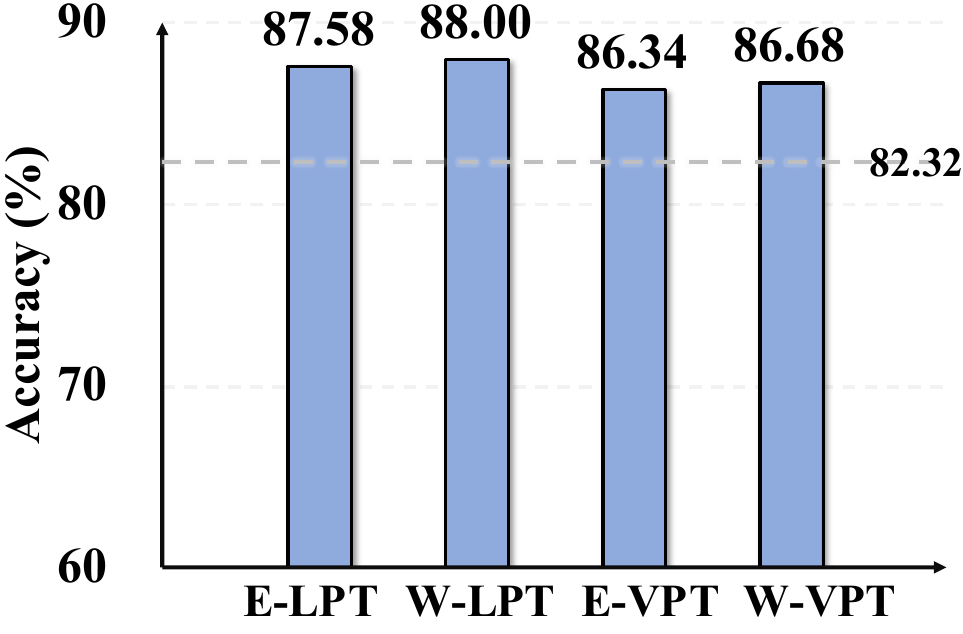}
				\caption{\officehome{}}
		\label{fig:officeagg}
	\end{subfigure}
    \end{minipage}
	\captionsetup{}
   
	\caption{Effects of different \textbf{Aggregation Ways}. E is short for Equal, while W is short for Weighted. The dotted line represents the effect of \clip{} with zero-shot inference. See details in \cref{sec:robustness}.}

	\label{fig:settingaggregation}

\end{figure}

To comprehensively evaluate the robustness of FPL, we conduct extensive experiments under different federated learning settings and prompt configurations, analyzing their impact on model performance. The following subsections detail our findings across communication epochs, aggregation strategies, client scales, and prompt lengths.

\noindent \textbf{Communication Epochs.} We assess the effect of varying the number of communication epochs on performance by conducting experiments on \cifarhun{} and \officehome{}. As shown in \cref{fig:settingrounds}, increasing the number of communication rounds leads to only marginal improvements in accuracy, while still maintaining a performance advantage over direct zero-shot inference of \clip{}. This suggests that Federated Prompt Learning is relatively robust to the number of training epochs, indicating that beyond a certain point, additional training contributes little to accuracy gains.

From a practical standpoint, this observation highlights an important trade-off: while increasing training rounds may slightly refine the model, it also incurs additional computational and communication costs. Therefore, in real-world applications, it is crucial to determine an optimal balance between training cost and performance improvement.

%%%%%%%%%% IVLPT 大表 %%%%%%%%%%
\begin{table*}[htbp]
\centering
\scriptsize{
\resizebox{\linewidth}{!}{
		\renewcommand\arraystretch{1.1}
\begin{tabular}{c||ccccIcIcccIcIccccccIc}
\hline\thickhline
\rowcolor{lightgray}
% \rowcolor{gray!10}
& \multicolumn{5}{cI}{Office-Home} &\multicolumn{4}{cI}{Office31} &\multicolumn{7}{c}{Domainnet}\\
\cline{2-17}
\rowcolor{lightgray}
% \rowcolor{gray!10}
\multirow{-2}{*}{Methods} & \textsl{A}  & \textsl{C}  & \textsl{PR} & \textsl{RW} & \textsl{AVG}
& \textsl{AM} & \textsl{D} & \textsl{W} & \textsl{AVG} & \textsl{C} & \textsl{I} &\textsl{P} &\textsl{Q} &\textsl{R} &\textsl{S} & \textsl{AVG}  \\
\hline
ZS-CLIP 
& 84.30 & 66.17 & 89.18 & 89.66 & 82.32
& 80.96 & 72.45 & 74.05 & 75.82 
& 87.69 & 69.66 & 79.77 & 28.11 & 91.91 & 84.82 & 73.66 \\
\hline\hline
\rowcolor{gray!5}
\multicolumn{17}{l}{\textsl{Lable skew: }$\beta=0.5$} \\
\cdashline{1-17}
VPT
& 85.08 & 74.89 & 93.21 & 92.25 & 86.62
& 85.66 & 90.82 & 89.62 & 88.70
&88.86 & 74.59 & 83.41 & 49.70 & 93.25 & 86.74 & 79.43\\

LPT
& 86.45 & 75.99 & 93.94 & 93.17 & 87.39
& 88.15 & 93.47 & 94.43 & 92.02
& 88.89 & 73.75 & 84.89 & 43.74 & 93.64 & 87.00 & 78.65\\

\rowcolor[HTML]{D7F6FF}
VLPT
& 85.95 & 78.19 & 95.22 & 93.01 & \textbf{88.09}\redorangeup{\textbf{5.77}}
& 87.94 & 94.90 & 95.44 & \textbf{92.76}\redorangeup{\textbf{16.94}}
& 90.41 & 74.52 & 84.78 & 53.98 & 93.65 & 87.82 & \textbf{80.86}\redorangeup{\textbf{7.20}}\\

\hline\hline
\rowcolor{gray!5}
\multicolumn{17}{l}{\textsl{Lable skew: }$\beta=1.0$} \\
\cdashline{1-17}
VPT
& 84.92 & 73.78 & 93.78 & 91.93 & 86.10
& 85.98 & 90.20 & 89.49 & 88.56
& 88.95 & 75.27 & 82.51 & 49.80 & 93.35 & 86.61 & 79.41\\

LPT
& 87.44 & 74.01 & 94.66 & 92.53 & 87.16
& 87.97 & 91.63 & 95.44 & 91.68
& 89.74 & 75.32 & 83.41 & 45.94 & 93.76 & 87.03 & 79.20\\

\rowcolor[HTML]{D7F6FF}
VLPT
& 87.07 & 77.34 & 94.99 & 92.53 & \textbf{87.98}\redorangeup{\textbf{5.66}}
& 87.19 & 94.29 & 95.06 & \textbf{92.18}\redorangeup{\textbf{16.36}}
& 91.40 & 73.94 & 84.70 & 55.12 & 93.46 & 87.10 & \textbf{80.95}\redorangeup{\textbf{7.29}}\\

\end{tabular}}}

\captionsetup{font=small}
\caption{ \small{Performance of different prompt learning in complex scenarios containing both \textbf{Label Skew and Domain Shift}. Please refer to \cref{sec:complex}
}}
\label{tab:compare_vlpt}

\end{table*}

\par
\noindent \textbf{Aggregation Ways.}
To investigate the impact of aggregation strategies in FL settings, we compare two common methods: Weighted and Equal. While Weighted aggregates client models by assigning weights based on the size of each client’s dataset, Equal assumes the weights of clients are equal.
\par
Our experiments on \cifarhun{} and \officehome{} shown in \cref{fig:settingaggregation} reveal an interesting pattern:
In label skew scenarios, \textbf{equal aggregation} leads to better performance, as treating clients uniformly prevents bias toward label-rich participants and ensures balanced optimization across classes.
In domain shift scenarios, \textbf{weighted aggregation} performs better, as emphasizing larger domains helps the global model capture the dominant distribution while still retaining cross-domain knowledge.
However, it is important to note that the overall difference in performance between these aggregation methods is \textbf{relatively small}. More importantly, both methods still outperform \clip{}’s zero-shot inference, reinforcing the robustness of federated prompt learning to different aggregation strategies. ggests that while aggregation choice may slightly influence results, FPL remains stable and effective across different settings.

%%%%%%%%%% client参数 %%%%%%%%%%
\begin {figure}[htbp]
	\centering
	\begin{minipage}{0.49\linewidth}
		\begin{subfigure}[b]{\textwidth}
		\includegraphics[width=\textwidth]{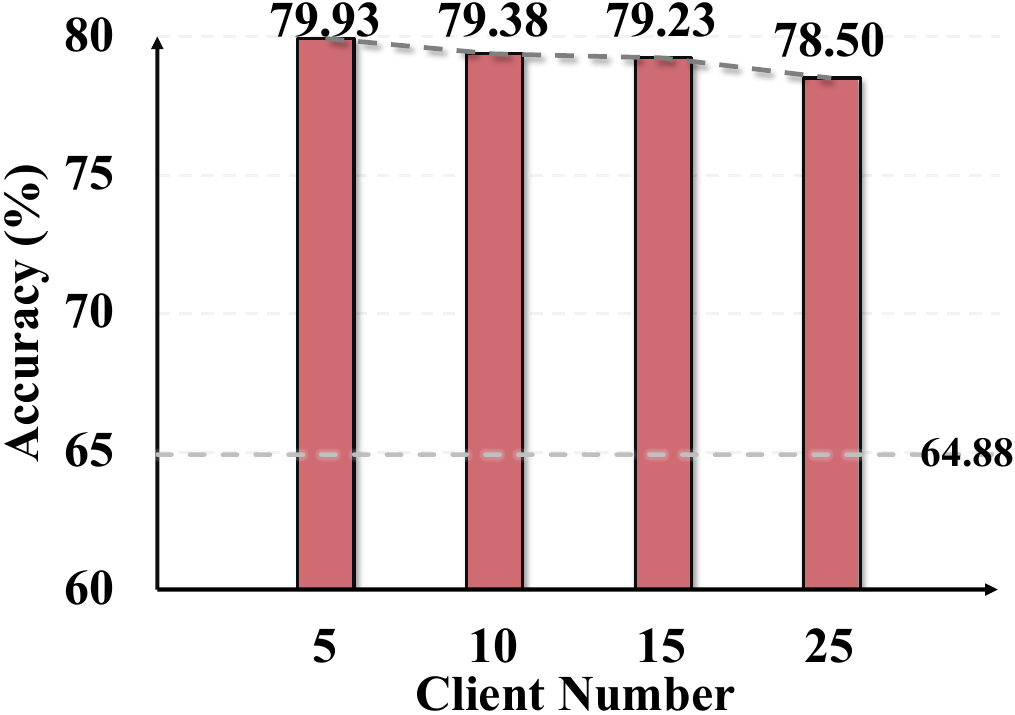}
				\caption{ \cifarhun{}}
		\label{fig:cifarclient}
	\end{subfigure}
    \end{minipage}
	\begin{minipage}{0.49\linewidth}
		\begin{subfigure}[b]{\textwidth}
		\includegraphics[width=\textwidth]{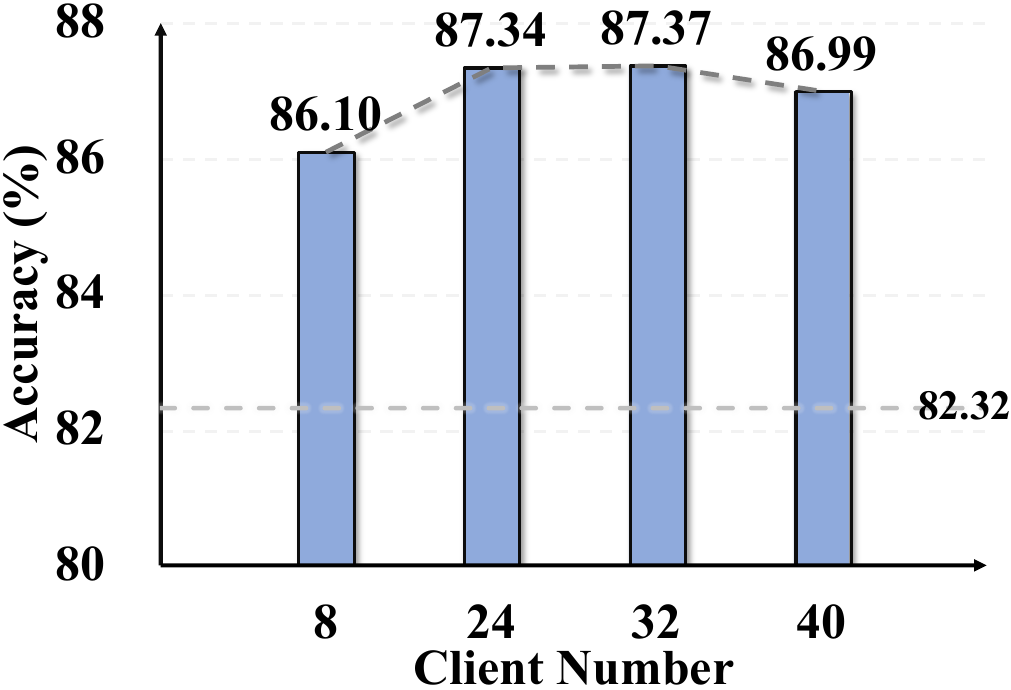}
				\caption{\officehome{}}
		\label{fig:officeclient}
	\end{subfigure}
    \end{minipage}

	\captionsetup{}

	\caption{Effects of different \textbf{Scale of Clients}. The dotted line represents the effect of \clip{} with zero-shot inference. Please refer to \cref{sec:robustness}.}

	\label{fig:settingclient}

\end{figure}

\par
\noindent \textbf{Scale of Clients.}
To evaluate the impact of client scale in federated learning settings, we conduct experiments on \cifarhun{} under label skew conditions and \domainnet{} under domain shift conditions. The results are presented in \cref{fig:settingclient}.
\par
Our findings indicate that in label skew scenarios, model accuracy gradually decreases as the number of clients increases. This is likely due to the increasing divergence in class distributions across clients, making it more challenging to learn a globally consistent prompt representation. In contrast, under domain shift scenarios, there is no clear pattern in accuracy fluctuations as the client count changes.
Nevertheless, across both settings, the variation in client scale does not significantly degrade model performance, and in all cases, FPL consistently outperforms \clip{}’s zero-shot inference. This demonstrates the robustness of FPL to different client scales, making it a reliable approach even in privacy-friendly environments with varying numbers of participants.

%%%%%%%%%% Length 大表 %%%%%%%%%%
\begin{table}[htbp]\small
\centering
\scriptsize{
\resizebox{\linewidth}{!}{
		\renewcommand\arraystretch{1.0}
\begin{tabular}{r||cccc}
\hline\thickhline
\rowcolor{lightgray}
& \multicolumn{4}{c}{Cifar100}\\
\cline{2-5}
\rowcolor{lightgray}
\multirow{-2}{*}{Length} & $\beta=0.3$ & $\beta=0.5$ & $\beta=1.0$ & $\beta=5.0$\\
\hline
\hline
ZS-CLIP & \multicolumn{4}{c}{64.88} \\
\cdashline{1-5}
\rowcolor{gray!10}
16
& 78.74\redorangeup{13.86} & 79.44\redorangeup{14.56} & 79.38\redorangeup{14.50} & 79.76\redorangeup{14.88}\\
32
& 79.95\redorangeup{15.07} & 79.87\redorangeup{14.99} & 80.49\redorangeup{15.61} & 80.32\redorangeup{15.44}\\

\rowcolor{gray!10}
64
& 80.01\redorangeup{15.13} & 80.10\redorangeup{15.22} & 80.58\redorangeup{15.70} & 80.92\redorangeup{16.04}\\
128
& \textbf{80.32}\redorangeup{15.44} & \textbf{80.37}\redorangeup{15.49} & \textbf{81.21}\redorangeup{16.33} & \textbf{81.56}\redorangeup{16.68} \\
\hline\hline\thickhline
\rowcolor{lightgray}
& \multicolumn{4}{c}{Office-Home}\\
\cline{2-5}
\rowcolor{lightgray}
\multirow{-2}{*}{Length} & $\beta=0.3$ & $\beta=0.5$ & $\beta=1.0$ & $\beta=5.0$\\
\hline
\hline
ZS-CLIP & \multicolumn{4}{c}{82.32} \\
% \hline
\cdashline{1-5}
\rowcolor{gray!10}
16
& 86.11\redorangeup{3.79} & 86.36\redorangeup{4.04} & 86.10\redorangeup{3.78} & 86.71\redorangeup{4.39} \\
32
& 86.09\redorangeup{3.77} & 86.38\redorangeup{4.06} & \textbf{86.47}\redorangeup{4.15} & 86.48\redorangeup{4.16} \\

\rowcolor{gray!10}
64
& 86.47\redorangeup{4.15} & 86.52\redorangeup{4.20} & 86.36\redorangeup{4.04} & 86.54\redorangeup{4.22} \\
128
& \textbf{86.62}\redorangeup{4.30} & \textbf{86.74}\redorangeup{4.42} & 86.22\redorangeup{3.90} & \textbf{86.85}\redorangeup{4.53}

\end{tabular}}}

\captionsetup{font=small}
\caption{\small{
Effects of different \textbf{Length of Prompt} on VPT in \cifarhun{} and \officehome{}. See details in \cref{sec:robustness}.
}}
 \label{tab:length}

\end{table}

\par
\noindent \textbf{Prompt Lengths.} We further examine the effect of prompt length as a critical parameter in prompt-based learning. We conduct experiments on \cifarhun{} and \officehome{}, using visual prompt tuning (VPT), and present the results in \cref{tab:length}. Key observations include: (a) Across multiple label skew settings, increasing the prompt length (e.g., doubling its size) generally leads to a slight improvement in accuracy.
(b) Even with minimal prompt tuning, FPL consistently surpasses \clip{}’s zero-shot inference, demonstrating its robustness to variations in prompt length.
While longer prompts can provide slight accuracy gains, they also increase the computational and memory overhead required for training. In real-world deployment, an optimal trade-off must be considered, balancing improved performance with resource constraints.

\par
The above experiments collectively demonstrate that changes in communication epochs, aggregation strategies, scale of clients, and prompt length lead to minor variations in performance. In all cases, FPL outperforms CLIP's zero-shot inference, reinforcing its adaptability and effectiveness in heterogeneous federated learning environments. We can answer \textbf{\textcolor{Q2green}{Q2}} in \cref{sec:intro} that: \textbf{\textcolor{Q2green}{A2}}: \textit{Federated Prompt Learning \textbf{remains robust} under different settings.}

\subsection{Collaboration Effect in Complex Scenarios}
\label{sec:complex}

In real-world environments, label skew and domain shift often coexist, creating highly complex learning conditions for federated prompt learning. To address this, we explore strategies to enhance Federated Prompt Learning (FPL) under such challenging conditions. Specifically, we investigate whether combining both vision prompt learning (VPT) and language prompt learning (LPT) can improve performance when computational resources allow. This combined approach, referred to as VLPT (Vision-Language Prompt Tuning), jointly leverages both textual and visual prompts to better align multimodal representations.

We evaluate VLPT across multiple datasets, including \officehome{}, \officeto{}, and \domainnet{}, where we introduce label skew on top of the existing domain shift settings. The experimental results are presented in \cref{tab:compare_vlpt} and several key findings are recapped below:
\begin{itemize}
    \item In all tested complex scenarios, VLPT consistently achieves the best performance, outperforming both standalone VPT and LPT across datasets.
    \item The results in \cref{sec:difference} suggest that LPT is more effective under domain shift, whereas VPT performs better in label skew scenarios. By combining both, VLPT benefits from their complementary strengths, improving the overall generalization capability.
    \item As seen in \cref{tab:compare_vlpt}, VLPT provides noticeable improvements across diverse domains, particularly in \domainnet{}, where domain shifts are more severe.
\end{itemize}

\par
\noindent \textbf{Practical Implications.}
As for \textbf{\textcolor{Q3purple}{Q3}} in \cref{sec:intro}, we have the conclusion that: \textbf{\textcolor{Q3purple}{A3}}: \textit{In real-world federated learning applications, where both label skew and domain shift are frequently common, VLPT provides an \textbf{effective solution for boosting} model performance.} If computational resources permit, utilizing both vision and language prompts simultaneously can significantly enhance the adaptability and overall robustness in federated learning settings.

\section{Conclusion and Future Work}
In this paper, we present a comprehensive empirical study on prompt learning for Vision Language Models (VLMs) in federated settings. Based on the experiments above, we can draw the key conclusions corresponding to the main research objectives as below:

\begin{itemize}
    \item {\protect \includegraphics[scale=0.08]{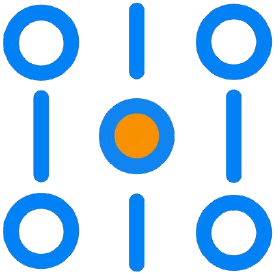}} \textbf{{\textcolor{Q1blue}{A1:}}}   \textit{Under label skew and domain shift, LPT and VPT show significant} \textbf{\textcolor{Q1blue}{Behavior Discrepancies}}. \textit{LPT excels in domain shift scenarios, while VPT works better in label skew since more similar updating directions are not always the “antidote” to data heterogeneity.}
  
  \item {\protect \includegraphics[scale=0.09]{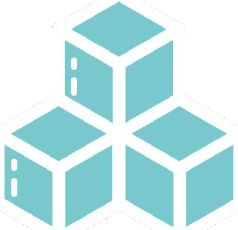}} \textbf{{\textcolor{Q2green}{A2:}}} \textit{Federated prompt learning exhibits} \textbf{\textcolor{Q2green}{Robustness Patterns}} \textit{when federated and prompt settings vary, such as communication epochs, aggregation ways, client scales and prompt lengths.}  
  
  \item {\protect\includegraphics[scale=0.08]{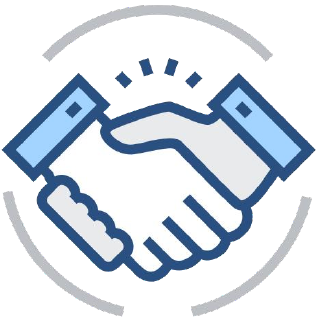}} \textbf{{\textcolor{Q3purple}{A3:}}} \textit{Integrating LPT and VPT helps to adapt in both branches of language and vision, thus showing} \textbf{\textcolor{Q3purple}{Collaboration Effect}} \textit{in complex scenarios containing label skew and domain shift.}
  
\end{itemize}

\par
While our findings highlight the feasibility and advantages of prompt learning in federated environments, there remain several avenues for future exploration. Current limitations in training time and computational resources restrict the scalability of our experiments, and further research is needed to enhance the efficiency, personalization, and adaptability of federated prompt learning. 

\appendix
\section*{Contribution Statement}
$^*$ Equal contribution. $^{\dagger}$ Corresponding author.

\clearpage
\newpage
\section*{Acknowledgments}
This work is supported by the National Key Research and Development Program of China (2024YFC3308400), National Natural Science Foundation of China under Grant (62032016, 62361166629, 62176188, 623B2080), Zhongguancun Laboratory, and the Wuhan University Undergraduate Innovation Research Fund Project.

{

\bibliographystyle{named}
\bibliography{ijcai25}

\begin{thebibliography}{}

\bibitem[\protect\citeauthoryear{Balakrishnan \bgroup \em et al.\egroup
  }{2019}]{Dirichlet}
Narayanaswamy Balakrishnan, Samuel Kotz, and Norman~L. Johnson.
\newblock Continuous multivariate distributions, volume 1: Models and
  applications.
\newblock 2019.

\bibitem[\protect\citeauthoryear{Bang \bgroup \em et al.\egroup
  }{2024}]{active_cvpr24}
Jihwan Bang, Sumyeong Ahn, and Jae-Gil Lee.
\newblock Active prompt learning in vision language models.
\newblock In {\em CVPR}, 2024.

\bibitem[\protect\citeauthoryear{Brown \bgroup \em et al.\egroup
  }{2020}]{gpt3_nips20}
Tom~B. Brown, Benjamin Mann, Nick Ryder, Melanie Subbiah, Jared Kaplan,
  Prafulla Dhariwal, Arvind Neelakantan, Pranav Shyam, Girish Sastry, Amanda
  Askell, Sandhini Agarwal, Ariel Herbert-Voss, Gretchen Krueger, Tom Henighan,
  Rewon Child, Aditya Ramesh, Daniel~M. Ziegler, Jeffrey Wu, Clemens Winter,
  Christopher Hesse, Mark Chen, Eric Sigler, Mateusz Litwin, Scott Gray,
  Benjamin Chess, Jack Clark, Christopher Berner, Sam McCandlish, Alec Radford,
  Ilya Sutskever, and Dario Amodei.
\newblock Language models are few-shot learners.
\newblock In {\em NeurIPS}, 2020.

\bibitem[\protect\citeauthoryear{Chen \bgroup \em et al.\egroup
  }{2023}]{PLOT_ICLR23}
Guangyi Chen, Weiran Yao, Xiangchen Song, Xinyue Li, Yongming Rao, and Kun
  Zhang.
\newblock {PLOT}: Prompt learning with optimal transport for vision-language
  models.
\newblock In {\em ICLR}, 2023.

\bibitem[\protect\citeauthoryear{Dai \bgroup \em et al.\egroup
  }{2023}]{Tackle_23}
Yutong Dai, Zeyuan Chen, Junnan Li, Shelby Heinecke, Lichao Sun, and Ran Xu.
\newblock Tackling data heterogeneity in federated learning with class
  prototypes.
\newblock In {\em AAAI}, 2023.

\bibitem[\protect\citeauthoryear{Feng \bgroup \em et al.\egroup
  }{2023}]{FedPR_CVPR23}
Chun-Mei Feng, Bangjun Li, Xinxing Xu, Yong Liu, Huazhu Fu, and Wangmeng Zuo.
\newblock Learning federated visual prompt in null space for mri
  reconstruction.
\newblock In {\em CVPR}, 2023.

\bibitem[\protect\citeauthoryear{Gu \bgroup \em et al.\egroup
  }{2023}]{SurveyofPromptVL_arXiv23}
Jindong Gu, Zhen Han, Shuo Chen, Ahmad Beirami, Bailan He, Gengyuan Zhang,
  Ruotong Liao, Yao Qin, Volker Tresp, and Philip Torr.
\newblock A systematic survey of prompt engineering on vision-language
  foundation models.
\newblock {\em arXiv preprint arXiv:2307.12980}, 2023.

\bibitem[\protect\citeauthoryear{Guo \bgroup \em et al.\egroup
  }{2023a}]{pFedPrompt_WWWW23}
Tao Guo, Song Guo, and Junxiao Wang.
\newblock Pfedprompt: Learning personalized prompt for vision-language models
  in federated learning.
\newblock In {\em WWW}, pages 1364--1374, 2023.

\bibitem[\protect\citeauthoryear{Guo \bgroup \em et al.\egroup
  }{2023b}]{PromptFL_TMC23}
Tao Guo, Song Guo, Junxiao Wang, Xueyang Tang, and Wenchao Xu.
\newblock Promptfl: Let federated participants cooperatively learn prompts
  instead of models-federated learning in age of foundation model.
\newblock {\em IEEE TMC}, 2023.

\bibitem[\protect\citeauthoryear{Hu \bgroup \em et al.\egroup
  }{2024}]{FedMut_24}
Ming Hu, Yue Cao, Anran Li, Zhiming Li, Chengwei Liu, Tianlin Li, Mingsong
  Chen, and Yang Liu.
\newblock Fedmut: Generalized federated learning via stochastic mutation.
\newblock In {\em AAAI}, 2024.

\bibitem[\protect\citeauthoryear{Huang \bgroup \em et al.\egroup
  }{2022}]{FCCL_22}
Wenke Huang, Mang Ye, and Bo~Du.
\newblock Learn from others and be yourself in heterogeneous federated
  learning.
\newblock In {\em CVPR}, 2022.

\bibitem[\protect\citeauthoryear{Huang \bgroup \em et al.\egroup
  }{2023a}]{FCCLPlus_TPAMI23}
Wenke Huang, Mang Ye, Zekun Shi, and Bo~Du.
\newblock Generalizable heterogeneous federated cross-correlation and instance
  similarity learning.
\newblock {\em TPAMI}, 2023.

\bibitem[\protect\citeauthoryear{Huang \bgroup \em et al.\egroup
  }{2023b}]{FPL_CVPR23}
Wenke Huang, Mang Ye, Zekun Shi, He~Li, and Bo~Du.
\newblock Rethinking federated learning with domain shift: A prototype view.
\newblock In {\em CVPR}, pages 16312--16322, 2023.

\bibitem[\protect\citeauthoryear{Huang \bgroup \em et al.\egroup
  }{2024}]{FLSurveyandBenchmarkforGenRobFair_PAMI24}
Wenke Huang, Mang Ye, Zekun Shi, Guancheng Wan, He~Li, Bo~Du, and Qiang Yang.
\newblock Federated learning for generalization, robustness, fairness: A survey
  and benchmark.
\newblock {\em IEEE PAMI}, 2024.

\bibitem[\protect\citeauthoryear{Jia \bgroup \em et al.\egroup
  }{2022}]{VPT_ECCV22}
Menglin Jia, Luming Tang, Bor-Chun Chen, Claire Cardie, Serge Belongie, Bharath
  Hariharan, and Ser-Nam Lim.
\newblock Visual prompt tuning.
\newblock In {\em ECCV}, 2022.

\bibitem[\protect\citeauthoryear{Kairouz \bgroup \em et al.\egroup
  }{2019}]{Advances_arXiv19}
Peter Kairouz, H~Brendan McMahan, Brendan Avent, Aur{\'e}lien Bellet, Mehdi
  Bennis, Arjun~Nitin Bhagoji, Kallista Bonawitz, Zachary Charles, Graham
  Cormode, Rachel Cummings, et~al.
\newblock Advances and open problems in federated learning.
\newblock {\em arXiv preprint arXiv:1912.04977}, 2019.

\bibitem[\protect\citeauthoryear{khattak \bgroup \em et al.\egroup
  }{2023a}]{MaPLe_cvpr23}
Muhammad~Uzair khattak, Hanoona Rasheed, Muhammad Maaz, Salman Khan, and
  Fahad~Shahbaz Khan.
\newblock Maple: Multi-modal prompt learning.
\newblock In {\em CVPR}, 2023.

\bibitem[\protect\citeauthoryear{Khattak \bgroup \em et al.\egroup
  }{2023b}]{PromptSRC_ICCV23}
Muhammad~Uzair Khattak, Syed~Talal Wasim, Muzammal Naseer, Salman Khan,
  Ming-Hsuan Yang, and Fahad~Shahbaz Khan.
\newblock Self-regulating prompts: Foundational model adaptation without
  forgetting.
\newblock In {\em ICCV}, 2023.

\bibitem[\protect\citeauthoryear{Kone{\v{c}}n{\'y} \bgroup \em et al.\egroup
  }{2016}]{FedOptimization_16}
Jakub Kone{\v{c}}n{\'y}, H.~Brendan McMahan, Daniel Ramage, and Peter
  Richt{\'{a}}rik.
\newblock Federated optimization: Distributed machine learning for on-device
  intelligence.
\newblock {\em CoRR}, abs/1610.02527, 2016.

\bibitem[\protect\citeauthoryear{Krizhevsky and Hinton}{2009}]{cifar_Toronto09}
A.~Krizhevsky and G.~Hinton.
\newblock Learning multiple layers of features from tiny images.
\newblock {\em Master's thesis, Department of Computer Science, University of
  Toronto}, 2009.

\bibitem[\protect\citeauthoryear{Lester \bgroup \em et al.\egroup
  }{2021}]{Prompttuning_emnlp21}
Brian Lester, Rami Al-Rfou, and Noah Constant.
\newblock The power of scale for parameter-efficient prompt tuning.
\newblock In {\em EMNLP}, 2021.

\bibitem[\protect\citeauthoryear{Li \bgroup \em et al.\egroup
  }{2020a}]{FLChallengesMethodsDirection_SPM20}
Tian Li, Anit~Kumar Sahu, Ameet Talwalkar, and Virginia Smith.
\newblock Federated learning: Challenges, methods, and future directions.
\newblock {\em IEEE Signal Process Mag}, pages 50--60, 2020.

\bibitem[\protect\citeauthoryear{Li \bgroup \em et al.\egroup
  }{2020b}]{Challenges_20}
Tian Li, Anit~Kumar Sahu, Ameet Talwalkar, and Virginia Smith.
\newblock Federated learning: Challenges, methods, and future directions.
\newblock {\em IEEE Signal Process Mag}, 2020.

\bibitem[\protect\citeauthoryear{Li \bgroup \em et al.\egroup
  }{2020c}]{FedProx_20}
Tian Li, Anit~Kumar Sahu, Manzil Zaheer, Maziar Sanjabi, Ameet Talwalkar, and
  Virginia Smith.
\newblock Federated optimization in heterogeneous networks.
\newblock In {\em Proceedings of Machine Learning and Systems}, 2020.

\bibitem[\protect\citeauthoryear{Li \bgroup \em et al.\egroup
  }{2021a}]{MOON_21}
Qinbin Li, Bingsheng He, and Dawn Song.
\newblock Model-contrastive federated learning.
\newblock In {\em CVPR}, 2021.

\bibitem[\protect\citeauthoryear{Li \bgroup \em et al.\egroup
  }{2021b}]{FedBN_ICLR21}
Xiaoxiao Li, Meirui Jiang, Xiaofei Zhang, Michael Kamp, and Qi~Dou.
\newblock Fed{\{}bn{\}}: Federated learning on non-{\{}iid{\}} features via
  local batch normalization.
\newblock In {\em ICLR}, 2021.

\bibitem[\protect\citeauthoryear{Li \bgroup \em et al.\egroup
  }{2022}]{Non-IIDStudy_22}
Qinbin Li, Yiqun Diao, Quan Chen, and Bingsheng He.
\newblock Federated learning on non-iid data silos: An experimental study.
\newblock In {\em ICDE}, 2022.

\bibitem[\protect\citeauthoryear{Li \bgroup \em et al.\egroup
  }{2023}]{VPwithDP_ICCV23}
Yizhe Li, Yu-Lin Tsai, Chia-Mu Yu, Pin-Yu Chen, and Xuebin Ren.
\newblock Exploring the benefits of visual prompting in differential privacy.
\newblock In {\em ICCV}, pages 5158--5167, 2023.

\bibitem[\protect\citeauthoryear{Li \bgroup \em et al.\egroup
  }{2024a}]{FedOTP_CVPR24}
Hongxia Li, Wei Huang, Jingya Wang, and Ye~Shi.
\newblock Global and local prompts cooperation via optimal transport for
  federated learning.
\newblock In {\em CVPR}, 2024.

\bibitem[\protect\citeauthoryear{Li \bgroup \em et al.\egroup
  }{2024b}]{promptkd_CVPR24}
Zheng Li, Xiang Li, Xinyi Fu, Xin Zhang, Weiqiang Wang, Shuo Chen, and Jian
  Yang.
\newblock Promptkd: Unsupervised prompt distillation for vision-language
  models.
\newblock In {\em CVPR}, 2024.

\bibitem[\protect\citeauthoryear{Liu \bgroup \em et al.\egroup }{2020}]{MFL_20}
Wei Liu, Li~Chen, Yunfei Chen, and Wenyi Zhang.
\newblock Accelerating federated learning via momentum gradient descent.
\newblock {\em IEEE TPDS}, 2020.

\bibitem[\protect\citeauthoryear{Lu \bgroup \em et al.\egroup
  }{2023}]{FedCLIP_DEB23}
Wang Lu, Xixu Hu, Jindong Wang, and Xing Xie.
\newblock Fedclip: Fast generalization and personalization for clip in
  federated learning.
\newblock {\em IEEE DEB}, 2023.

\bibitem[\protect\citeauthoryear{Luo \bgroup \em et al.\egroup
  }{2021}]{CCVR_NeurIPS21}
Mi~Luo, Fei Chen, Dapeng Hu, Yifan Zhang, Jian Liang, and Jiashi Feng.
\newblock No fear of heterogeneity: Classifier calibration for federated
  learning with non-iid data.
\newblock In {\em NeurIPS}, 2021.

\bibitem[\protect\citeauthoryear{Ma \bgroup \em et al.\egroup
  }{2022}]{SOTAIIDSurvey_22}
Xiaodong Ma, Jia Zhu, Zhihao Lin, Shanxuan Chen, and Yangjie Qin.
\newblock A state-of-the-art survey on solving non-iid data in federated
  learning.
\newblock {\em Future Gener Comput Syst}, 2022.

\bibitem[\protect\citeauthoryear{McMahan \bgroup \em et al.\egroup
  }{2017}]{FedAvg_AISTATS17}
Brendan McMahan, Eider Moore, Daniel Ramage, Seth Hampson, and Blaise~Aguera
  y~Arcas.
\newblock Communication-efficient learning of deep networks from decentralized
  data.
\newblock In {\em AISTATS}, pages 1273--1282, 2017.

\bibitem[\protect\citeauthoryear{Oord \bgroup \em et al.\egroup
  }{2018}]{infonce_arXiv18}
Aaron van~den Oord, Yazhe Li, and Oriol Vinyals.
\newblock Representation learning with contrastive predictive coding.
\newblock {\em arXiv preprint arXiv:1807.03748}, 2018.

\bibitem[\protect\citeauthoryear{Peng \bgroup \em et al.\egroup
  }{2019}]{DomainNet_ICCV19}
Xingchao Peng, Qinxun Bai, Xide Xia, Zijun Huang, Kate Saenko, and Bo~Wang.
\newblock Moment matching for multi-source domain adaptation.
\newblock In {\em ICCV}, pages 1406--1415, 2019.

\bibitem[\protect\citeauthoryear{Qiu \bgroup \em et al.\egroup
  }{2024}]{FedTPG_ICLR24}
Chen Qiu, Xingyu Li, Chaithanya~Kumar Mummadi, Madan~Ravi Ganesh, Zhenzhen Li,
  Lu~Peng, and Wan-Yi Lin.
\newblock Federated text-driven prompt generation for vision-language models.
\newblock In {\em ICLR}, 2024.

\bibitem[\protect\citeauthoryear{Radford \bgroup \em et al.\egroup
  }{2021}]{CLIP_21}
Alec Radford, Jong~Wook Kim, Chris Hallacy, Aditya Ramesh, Gabriel Goh,
  Sandhini Agarwal, Girish Sastry, Amanda Askell, Pamela Mishkin, Jack Clark,
  Gretchen Krueger, and Ilya Sutskever.
\newblock Learning transferable visual models from natural language
  supervision.
\newblock In {\em ICML}, pages 8748--8763, 2021.

\bibitem[\protect\citeauthoryear{Saenko \bgroup \em et al.\egroup
  }{2010}]{office31_ECCV10}
Kate Saenko, Brian Kulis, Mario Fritz, and Trevor Darrell.
\newblock Adapting visual category models to new domains.
\newblock In {\em ECCV}, pages 213--226, 2010.

\bibitem[\protect\citeauthoryear{Sattler \bgroup \em et al.\egroup
  }{2021}]{CFL_21}
Felix Sattler, Klaus-Robert Müller, and Wojciech Samek.
\newblock Clustered federated learning: Model-agnostic distributed multitask
  optimization under privacy constraints.
\newblock {\em IEEE TNNLS}, 2021.

\bibitem[\protect\citeauthoryear{Shin \bgroup \em et al.\egroup
  }{2020}]{AutoPrompt_EMNLP20}
Taylor Shin, Yasaman Razeghi, Robert~L Logan~IV, Eric Wallace, and Sameer
  Singh.
\newblock Autoprompt: Eliciting knowledge from language models with
  automatically generated prompts.
\newblock In {\em EMNLP}, 2020.

\bibitem[\protect\citeauthoryear{Su \bgroup \em et al.\egroup
  }{2024}]{FedAPT_AAAI24}
Shangchao Su, Mingzhao Yang, Bin Li, and Xiangyang Xue.
\newblock Federated adaptive prompt tuning for multi-domain collaborative
  learning.
\newblock In {\em AAAI}, pages 15117--15125, 2024.

\bibitem[\protect\citeauthoryear{Tan \bgroup \em et al.\egroup
  }{2023}]{FedStar_AAAI23}
Yue Tan, Yixin Liu, Guodong Long, Jing Jiang, Qinghua Lu, and Chengqi Zhang.
\newblock Federated learning on non-iid graphs via structural knowledge
  sharing.
\newblock In {\em AAAI}, 2023.

\bibitem[\protect\citeauthoryear{Venkateswara \bgroup \em et al.\egroup
  }{2017}]{officehome_CVPR17}
Hemanth Venkateswara, Jose Eusebio, Shayok Chakraborty, and Sethuraman
  Panchanathan.
\newblock Deep hashing network for unsupervised domain adaptation.
\newblock In {\em CVPR}, pages 5018--5027, 2017.

\bibitem[\protect\citeauthoryear{Wang \bgroup \em et al.\egroup
  }{2025}]{FREIB_AAAI2025}
Zhihao Wang, He~Bai, Wenke Huang, Duantengchuan Li, Jian Wang, and Bing Li.
\newblock Federated recommendation with explicitly encoding item bias.
\newblock In {\em AAAI}, 2025.

\bibitem[\protect\citeauthoryear{Wei \bgroup \em et al.\egroup
  }{2023}]{FedDPT_arXiv23}
Guoyizhe Wei, Feng Wang, Anshul Shah, and Rama Chellappa.
\newblock Dual prompt tuning for domain-aware federated learning.
\newblock {\em arXiv preprint arXiv:2310.03103}, 2023.

\bibitem[\protect\citeauthoryear{Yang \bgroup \em et al.\egroup
  }{2023}]{pFedPG_ICCV23}
Fu-En Yang, Chien-Yi Wang, and Yu-Chiang~Frank Wang.
\newblock Efficient model personalization in federated learning via
  client-specific prompt generation.
\newblock In {\em ICCV}, pages 19159--19168, 2023.

\bibitem[\protect\citeauthoryear{Yao \bgroup \em et al.\egroup
  }{2023}]{KgCoOp_cvpr23}
Hantao Yao, Rui Zhang, and Changsheng Xu.
\newblock Visual-language prompt tuning with knowledge-guided context
  optimization.
\newblock In {\em CVPR}, 2023.

\bibitem[\protect\citeauthoryear{Ye \bgroup \em et al.\egroup
  }{2023}]{Heterogeneous_Survey_2023}
Mang Ye, Xiuwen Fang, Bo~Du, Pong~C. Yuen, and Dacheng Tao.
\newblock Heterogeneous federated learning: State-of-the-art and research
  challenges.
\newblock {\em ACM Comput. Surv.}, 2023.

\bibitem[\protect\citeauthoryear{Zhao \bgroup \em et al.\egroup
  }{2018}]{FLwithNonIID_arXiv18}
Yue Zhao, Meng Li, Liangzhen Lai, Naveen Suda, Damon Civin, and Vikas Chandra.
\newblock Federated learning with non-iid data.
\newblock {\em arXiv preprint arXiv:1806.00582}, 2018.

\bibitem[\protect\citeauthoryear{Zhou \bgroup \em et al.\egroup
  }{2022}]{CoOp_IJCV22}
Kaiyang Zhou, Jingkang Yang, Chen~Change Loy, and Ziwei Liu.
\newblock Learning to prompt for vision-language models.
\newblock {\em IJCV}, 130(9):2337--2348, 2022.

\end{thebibliography}
}
\end{document}